\documentclass[11pt]{article}

\usepackage[final]{acl}

\usepackage{times}
\usepackage{latexsym}

\usepackage[T1]{fontenc}

\usepackage[utf8]{inputenc}

\usepackage{microtype}

\usepackage{inconsolata}

\usepackage{graphicx}

\usepackage{amsmath}
\usepackage{amsfonts}
\usepackage{booktabs}
\usepackage[table]{xcolor}
\usepackage{pifont}
\usepackage{enumitem}
\usepackage{dsfont}
\usepackage{multirow}
\usepackage{hyperref}
\usepackage{tcolorbox}
\tcbuselibrary{listings, breakable}
\usepackage{listings}
\usepackage{dashrule}
\tcbuselibrary{skins}
\usepackage{geometry}
\usepackage{tikz}

\newcommand{\cmark}{\ding{51}}
\newcommand{\xmark}{\ding{55}}

\definecolor{motivationcolor}{RGB}{145, 168, 188}
\definecolor{lightmotivation}{RGB}{237, 243, 248}
\definecolor{reasoningcolor}{RGB}{147, 182, 165}
\definecolor{lightreasoning}{RGB}{238, 247, 242}
\definecolor{methodcolor}{RGB}{172, 152, 179}
\definecolor{lightmethod}{RGB}{245, 241, 247}
\definecolor{highlightcolor}{RGB}{198, 134, 128}

\newtcolorbox{motivationbox}[1][]{  
    enhanced,  
    colback=lightmotivation,  
    colframe=motivationcolor,  
    fonttitle=\bfseries\large,  
    title={\textcolor{white}{Motivation}},  
    coltitle=white,  
    attach boxed title to top left={yshift=-2mm, xshift=5mm},  
    boxed title style={colback=motivationcolor, sharp corners},  
    sharp corners,  
    boxrule=0.8pt,  
    left=5pt, right=5pt, top=5pt, bottom=5pt,  
    fontupper=\small,  
    #1  
}

\newtcolorbox{reasoningbox}[1][]{  
    enhanced,  
    colback=lightreasoning,  
    colframe=reasoningcolor,  
    fonttitle=\bfseries\large,  
    title={\textcolor{white}{Reasoning (Representative Excerpts)}},  
    coltitle=white,  
    attach boxed title to top left={yshift=-2mm, xshift=5mm},  
    boxed title style={colback=reasoningcolor, sharp corners},  
    sharp corners,  
    boxrule=0.8pt,  
    left=5pt, right=5pt, top=5pt, bottom=5pt,  
    fontupper=\small,  
    #1  
}

\newtcolorbox{methodbox}[1][]{  
    enhanced,  
    colback=lightmethod,  
    colframe=methodcolor,  
    fonttitle=\bfseries\large,  
    title={\textcolor{white}{Method}},  
    coltitle=white,  
    attach boxed title to top left={yshift=-2mm, xshift=5mm},  
    boxed title style={colback=methodcolor, sharp corners},  
    sharp corners,  
    boxrule=0.8pt,  
    left=5pt, right=5pt, top=5pt, bottom=5pt,  
    fontupper=\small,  
    #1  
}

\definecolor{baselinecolor}{RGB}{210, 160, 135}
\definecolor{lightbaseline}{RGB}{252, 246, 242}

\newtcolorbox{baselinebox}[2][]{  
    enhanced,  
    colback=lightbaseline,  
    colframe=baselinecolor,  
    fonttitle=\bfseries\large,  
    title={\textcolor{white}{#2}},  
    coltitle=white,  
    attach boxed title to top left={yshift=-2mm, xshift=5mm},  
    boxed title style={colback=baselinecolor, sharp corners},  
    sharp corners,  
    boxrule=0.8pt,  
    left=5pt, right=5pt, top=5pt, bottom=5pt,  
    fontupper=\small,  
    #1  
}

\newcommand{\insight}[1]{\textbf{\textcolor{highlightcolor}{#1}}}

\title{MoRI: Learning Motivation-Grounded Reasoning for Scientific Ideation in Large Language Models}

\author{
  Chenyang Gu, Jiahao Cheng, Meicong Zhang, Pujun Zheng, Jinquan Zheng, Guoxiu He\thanks{\ \ Corresponding author.} \\
  School of Economics and Management, East China Normal University \\
  \texttt{\{cygu, chengjiahao, mczhang, pjzheng, jqzheng\}@stu.ecnu.edu.cn},\\
  \texttt{gxhe@fem.ecnu.edu.cn}
}

\begin{document}
\maketitle

\begin{abstract}
Scientific ideation aims to propose novel solutions within a given scientific context. Existing LLM-based agentic approaches emulate human research workflows, yet inadequately model scientific reasoning, resulting in surface-level conceptual recombinations that lack technical depth and scientific grounding. To address this issue, we propose \textbf{MoRI} (\textbf{Mo}tivation-grounded \textbf{R}easoning for Scientific \textbf{I}deation), a framework that enables LLMs to explicitly learn the reasoning process from research motivations to methodologies. The base LLM is initialized via supervised fine-tuning to generate a research motivation from a given context, and is subsequently trained under a composite reinforcement learning reward that approximates scientific rigor: (1) entropy-aware information gain encourages the model to uncover and elaborate high-complexity technical details grounded in ground-truth methodologies, and (2) contrastive semantic gain constrains the reasoning trajectory to remain conceptually aligned with scientifically valid solutions. Empirical results show that MoRI consistently outperforms strong commercial LLMs and complex agentic baselines across multiple dimensions, including novelty, technical rigor, and feasibility. The code is available on \href{https://github.com/ECNU-Text-Computing/IdeaGeneration}{GitHub}.  
\end{abstract}

\section{Introduction}
\label{sec:intro}
The pursuit of scientific discovery, from hypothesis formulation to experimental design, represents a highly complex form of human cognition. With the advent and improvement of Large Language Models (LLMs) \citep{radford2019language, achiam2023gpt, touvron2023llama, guo2025deepseek}, this landscape is undergoing a paradigm shift. 
While LLMs are evolving beyond general-purpose chatbots toward more capable scientific assistants and even researchers \citep{yao2022react, wang2024autosurvey, tang2025large, tang2025ai, lu2024ai}, the core challenge of \textbf{Scientific Ideation} (\textit{i.e.}, Scientific Idea Generation) remains largely unsolved. A high-quality scientific idea is not merely a fluent description of a method; it is a coherent logical structure that maps a specific \textit{Research Context} onto a feasible \textit{Methodology} through a principled \textit{Motivation} \citep{keya2025sci, lei2025motivgraph}.

\begin{figure*}[t]
    \centering
    \includegraphics[width=0.965\textwidth]{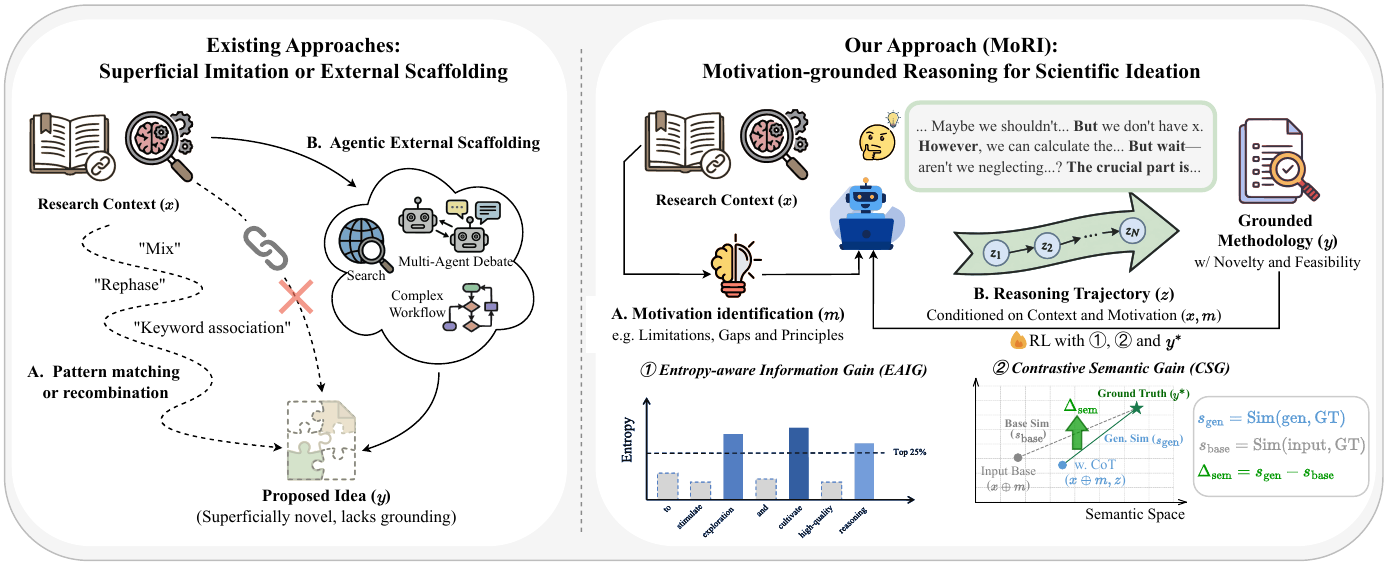}
    \caption{\textbf{Conceptual comparison.} 
    Unlike existing approaches that rely on pattern recombination or computationally expensive external scaffolding, MoRI internalizes scientific ideation through learning motivation-grounded reasoning. It initially identifies a Motivation ($m$) from a given Context ($x$), then generates a Reasoning Trajectory ($z$) to deduce a grounded Methodology ($y$), which is optimized via our composite RL rewards.
    }
    \label{fig:intro}
\end{figure*}

As shown in Figure~\ref{fig:intro}, current approaches to automating this process face significant limitations as native LLMs lack a deep understanding of underlying scientific motivations \citep{si2024can, schapiro2025combinatorial, kumar2025can}. To mitigate this, recent studies have employed agentic scaffolding, orchestrating LLMs within complex, iterative workflows \citep{su2025many, baek2025researchagent, li2025chain, yamada2025ai}. However, they largely rely on human-designed heuristics rather than enhancing intrinsic reasoning capabilities, resulting in conceptual recombinations that are superficially novel but lack substantive technical depth and scientific grounding. As noted in the "Bitter Lesson" \citep{sutton2019bitter}, relying on external scaffolds is often less scalable and effective than internalizing reasoning capabilities for high-quality ideation.

To bridge this gap, we propose a novel framework, \textbf{MoRI} (\textbf{Mo}tivation-grounded \textbf{R}easoning for Scientific \textbf{I}deation). Unlike prior approaches relying on external agentic scaffolding or human-designed iterative pipelines, MoRI seeks to internalize the reasoning process of scientific discovery. We formulate scientific ideation as a motivation-grounded \citep{lei2025motivgraph, becker2024computational} reasoning task: the model is initialized to identify a principled Motivation ($m$) from the Research Context ($x$), and then learns to generate a Reasoning Trajectory ($z$) that logically deduces a grounded Methodology ($y$). This formulation forces the model to move beyond context completion toward deliberate, motivation-driven problem solving. The conceptual comparison is shown in Figure~\ref{fig:intro}.

To instantiate this perspective, we curate an ideation dataset from accepted ICLR 2024–2025 papers, using the \textit{Method} sections as proxies for high-quality ideas. We first initialize the model via supervised fine-tuning (SFT) to establish foundational capabilities in motivation identification and reasoning generation \citep{wei2022chain, guo2025deepseek}. Building upon this, we optimize the model with reinforcement learning (RL). This stage is designed to internalize scientific reasoning, where the policy learns to construct logical reasoning trajectories that bridge the identified research motivation and the ground-truth methodology. At inference, MoRI mirrors this structure by first proposing a motivation and then performing motivation-conditioned reasoning to generate the final idea.

A fundamental challenge in applying RL to scientific ideation is the absence of deterministic verifiers that allow validation against standard solutions \citep{jin2025search, feng2025retool, shao2024deepseekmath}. To address this, MoRI introduces a novel reward framework serving as a surrogate for scientific rigor. We design \textbf{entropy-aware information gain} to incentivize the elucidation of high-complexity technical details in ground truth methodologies, while proposing \textbf{contrastive semantic gain} to ensure the reasoning aligns with the conceptual trajectory of the ground truth. This \textbf{synergy} effectively balances \textit{micro-level} technical depth with \textit{macro-level} logical direction, which enables the model to internalize professional standards of scientific inquiry, fostering ideas that are both technically grounded and conceptually innovative. To further stabilize optimization, we incorporate length-anchoring regularization and format checks to prevent reward-hacking behaviors, such as reasoning shortcuts or method leakage \citep{he2025thinkdial}.

We implement MoRI with DeepSeek-R1-Distilled-Qwen-14B \citep{guo2025deepseek} and compare it against leading commercial models and agentic baselines. To ensure rigorous assessment, we employ a hybrid evaluation protocol combining retrieval-augmented LLM judges and human experts. The high correlation between these metrics supports the reliability of our results. Experimental findings demonstrate that MoRI consistently generates scientific ideas with superior novelty, technical rigor, and feasibility.
Our main contributions are summarized as follows:

\begin{itemize}[leftmargin=*, noitemsep, topsep=0pt]
    \item We formulate scientific ideation as a reasoning-driven process that generates grounded methodologies from context via intermediate motivations, and construct an ideation dataset from recent ICLR papers to support model training.
    
    \item We propose \textbf{MoRI}, a framework that leverages RL to internalize the reasoning process bridging research motivations and methodologies, transcending simple imitation.

    \item We introduce a composite reward integrating \textbf{entropy-aware information gain} and \textbf{contrastive semantic gain}, whose synergy effectively guides scientific reasoning by balancing technical depth with conceptual direction.

    \item Extensive evaluations via both LLM and human judges demonstrate MoRI's consistent superiority in generating novel and feasible scientific ideas over strong baselines.
\end{itemize}

\section{Related Work}
\label{sec:related}

\subsection{LLMs for Scientific Ideation}

The automation of scientific discovery has progressed from early rule-based expert systems \citep{lindsay1980applications, langley1987scientific} and literature-based discovery algorithms \citep{swanson1986undiscovered, king2009automation} to generative paradigms enabled by LLMs. Within this domain, scientific ideation stands as a foundational upstream task. Current frameworks predominantly adopt agentic paradigms, orchestrating LLMs to simulate research workflows via iterative research, multi-agent debate, and autonomous peer review \citep{wang2024scimon, lu2024ai, baek2025researchagent, tang2025ai, yamada2025ai, weng2024cycleresearcher, su2025many}. 
However, these methods rely on external scaffolding rather than enhancing internal reasoning abilities. Without explicit scientific training, LLMs often struggle to model research motivations and reason from them to feasible methodologies, resulting in ideas that appear superficially novel but technically derivative \citep{si2024can, kumar2025can}.

In contrast to agentic approaches, we propose to internalize the ideation process through explicit scientific training. By employing motivation-conditioned RL to optimize the reasoning path bridging motivations and methodologies, our model learns professional scientific standards through reasoning rather than heuristic design.

\subsection{RL for Open-Ended Reasoning}

RL has proven effective in enhancing LLM reasoning \citep{wei2022chain, guo2025deepseek, jaech2024openai}, particularly in domains like mathematics and tool utilization where deterministic verifiers can guide long-horizon thought processes \citep{guo2025deepseek, shao2024deepseekmath, jin2025search, feng2025retool}. To generalize RL beyond deterministic settings and handle tasks with sparse or noisy signals, recent research has explored more robust reward frameworks. One direction investigates verifier-free approaches that leverage intrinsic signals as surrogate rewards \citep{liu2025nover, yu2025rlpr, zhou2025reinforcing, gurung2025learning}. These methods are closely related to our setting, but scientific ideation differs from tasks with relatively canonical reference answers: multiple long-form Method sections can be scientifically valid even when they diverge lexically, which makes direct reference-probability or perplexity rewards noisy in our setting. Concurrently, to accommodate open-ended scenarios, other works employ model-based evaluations via structured rubrics or preference models to approximate human judgment \citep{gunjal2025rubrics, shao2025dr, bhaskar2025language, zheng2023judging}. Furthermore, recent studies have introduced information-theoretic objectives, such as information gain or semantic diversity, to motivate exploration and content quality \citep{wang2025information, li2025jointly, liu2025breaking}.

However, these frameworks pose challenges for scientific ideation. Intrinsic signals can lack the rigor required for scientific inquiry, while model-based rewards are computationally expensive and prone to hacking \citep{zhao2025one}. MoRI addresses this by formulating a composite reward for scientific reasoning. Distinct from prior entropy-based methods that filter reasoning tokens \citep{wang2025beyond}, our \textit{entropy-aware information gain} targets high-complexity technical details in the ground truth, ensuring depth. Complemented by \textit{contrastive semantic gain} for directional alignment, this synergy enables the model to internalize professional standards, fostering robust ideation abilities.

\begin{figure*}[t]
    \centering
    \includegraphics[width=1.0\textwidth]{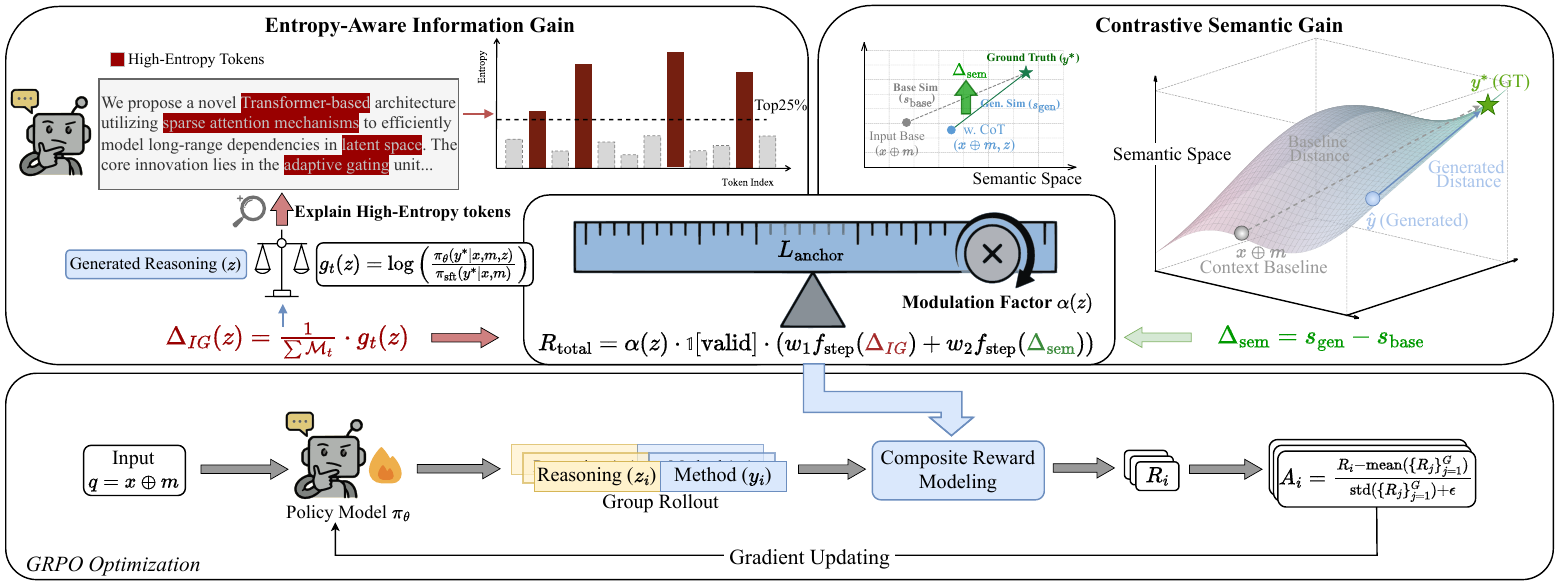}
    \caption{\textbf{Overview of MoRI.} Our framework optimizes reasoning via GRPO (Bottom) using composite rewards: \textbf{Entropy-Aware Information Gain} (Left) for high-entropy explanation and technical depth and \textbf{Contrastive Semantic Gain} (Right) for logical direction alignment, modulated by \textbf{Length Anchoring} (Center) to enforce reasoning depth.}
\label{fig:reward}
\end{figure*}

\section{Methodology}
\label{sec:methodology}

We propose \textbf{MoRI}, a framework that internalizes scientific ideation within LLMs. Building on an SFT-initialized model capable of generating motivations from a given context, MoRI further optimizes the reasoning that transforms motivations into solutions (Figure~\ref{fig:reward}). This process is governed by a composite reward mechanism where entropy-aware information gain and contrastive semantic gain synergize to ensure both technical depth and logical direction, strictly modulated by length anchoring and format constraints to enforce robust reasoning. Finally, inference follows a cascaded procedure to generate motivation-aligned methods.

\subsection{Problem Formulation}
\label{subsec:problem_formulation}

We formulate scientific ideation as a multi-stage conditional reasoning problem. Unlike approaches that map research contexts directly to solutions, we posit that high-quality ideas stem from a principled reasoning process driven by specific research motivations.

Let $x$ denote the research context which includes topic, background, and key references. Let $m$ represent the research motivation, which encompasses not only the identified gaps in prior work but also the scientific principles and high-level directions guiding the proposed solution. Let $y$ denote the detailed methodology. We introduce a reasoning trajectory $z$ that acts as the logical bridge connecting the motivation to the methodology. The ideation process is decomposed into two sequential policies:
\begin{equation}
\resizebox{0.89\hsize}{!}{$
\begin{aligned}
    x \xrightarrow[\text{Motivation Proposal}]{m \sim \pi_{\phi}(\cdot|x)} m \xrightarrow[\text{Ideation w/ Reasoning}]{(z, y) \sim \pi_{\theta}(\cdot|x, m)} (z, y)
\end{aligned}
$}
\label{eq:infer}
\end{equation}
which correspond to two stages of the same underlying model. $\pi_{\phi}$ is the motivation proposal policy and $\pi_{\theta}$ is the reasoning-driven ideation policy. During RL, we condition $\pi_{\theta}$ on the ground-truth motivation $m$ to optimize the reasoning trajectory $z$ and the method $y$. The objective is to enable the model to learn and generalize the cognitive pattern of deducing feasible solutions from identified motivations, moving beyond the surface-level imitation of context-method correlations.

\subsection{Data Construction and Initialization}
\label{subsec:data_construction}

To establish a foundational corpus for SFT initialization and motivation-grounded RL, we construct a dataset $\mathcal{D} = \{(x_i, m_i, z_i, y^*_i)\}_{i=1}^N$ derived from accepted ICLR 2024–2025 papers. The data pipeline employs LLMs to extract standardized research contexts $x_i$, motivations $m_i$, and de-symbolized method descriptions $y^*_i$ from raw PDF. To bridge the gap between context and method, we introduce a \textbf{posterior reconstruction strategy} that synthesizes the reasoning trajectories $z_i$ by reverse-engineering the logical path from ground-truth methods. 
We initialize our underlying model via SFT on a mixture of tasks, encouraging it both to generate motivations from context ($\pi_{\phi}$) and develop preliminary reasoning to produce methodologies ($\pi_{\theta}$) within the same model.
See Appendix~\ref{app:data_construction} for details.

\subsection{Motivation-Grounded RL} 
\label{subsec:rl_framework}

Moving beyond the imitation boundaries of naive SFT, we utilize RL to stimulate exploration and learn high-quality reasoning patterns specifically for motivation-grounded ideation. This transition shifts the learning objective from surface-level text mimicry to the internalization of robust scientific logic, enabling the model to derive valid methodologies from motivations.

We adopt Group Relative Policy Optimization (GRPO) \citep{shao2024deepseekmath} with token-level loss and clip-higher \citep{yu2025dapo} as our optimization algorithm. 
During training, the model receives the joint input $q = x \oplus m$ (Context concatenated with Motivation). For each input, the policy $\pi_{\theta}$ samples a group of outputs $\{o_1, \dots, o_G\}$, where each $o_i$ consists of a reasoning trajectory $z_i$ and the generated method content. The optimization objective is defined as (For brevity, the KL penalty term is omitted in this formulation.):
\begin{equation}
\resizebox{0.89\hsize}{!}{$
\begin{aligned}
    \mathcal{J}(\theta) = & \mathbb{E}_{q \sim \mathcal{D}, \{o_i\}_{i=1}^G \sim \pi_{\theta_{\text{old}}}} \bigg[ \frac{1}{\sum_{i=1}^G |o_i|} \sum_{i=1}^G  \sum_{t=1}^{|o_i|} \\
    & \min \left( r_{i,t} A_i, \text{clip}(r_{i,t}, 1-\varepsilon_\text{low}, 1+\varepsilon_\text{high}) A_i \right) \bigg]
\end{aligned}
$}
\end{equation}
where $r_{i,t}$ is the importance sampling ratio, and $A_i = \frac{R(o_i) - \text{mean}(\{R(o_j)\}_{j=1}^G)}{\text{std}(\{R(o_j)\}_{j=1}^G)+\epsilon}$ is the advantage computed via group normalization. 
Here, $R(o_i)$ is our composite reward function, which serves as the core signal to guide the reasoning optimization.

\subsection{Composite Reward Modeling}
\label{subsec:reward_modeling}

We introduce a Dual-Granularity Reward Framework that evaluates the generated reasoning from two complementary dimensions: \textit{Micro-level Explainability} and \textit{Macro-level Semantic Alignment}.

\noindent\textbf{Theoretical Motivation.} We posit that a high-quality reasoning trajectory $z$ serves a dual purpose. First, it should be \textbf{explanatory}: it must provide the necessary logical context to render the complex, high-information details of the ground-truth method $y^*$ more probable. Second, it should be \textbf{constructive}: the resulting method $y$ must represent a significant semantic advancement from the problem context $x$ towards the solution space, rather than a mere restatement of the input. Based on this, we design two reward components: Entropy-Aware Information Gain (EAIG) and Contrastive Semantic Gain (CSG).

\subsubsection{Entropy-Aware Information Gain}

To ensure the generated ideas are grounded in technical substance rather than superficial fluency, we design the Entropy-Aware Information Gain (EAIG). We employ a decoupled strategy that separates the identification of critical content from the evaluation of reasoning quality.

\noindent\textbf{Selection via Entropy.} Not all tokens in a scientific method contribute equally to innovation. Functional phrases such as "we propose to" are highly predictable and carry little scientific substance, whereas specific algorithms or parameters represent the core "hard knowledge." To target the latter, we utilize entropy as a filter. As shown in Figure~\ref{fig:reward} (Left), we compute the predictive entropy at each position of the ground-truth method under the fixed SFT model, conditioned on the full RL input $(x, m)$ and the teacher-forced prefix $y^*_{<t}$:

\begin{equation}
\begin{aligned}
H_t = - \sum_{v \in \mathcal{V}}
&\;\pi_{\mathrm{sft}}\!\left(v \mid x, m, y^{*}_{<t}\right) \\
&\; \cdot \log \pi_{\mathrm{sft}}\!\left(v \mid x, m, y^{*}_{<t}\right).
\end{aligned}
\end{equation}

A mask $\mathcal{M}_t$ is then generated to activate only for the top 25\% of tokens with the highest entropy, effectively isolating the high-complexity technical details.

We empirically validate that this threshold preferentially selects substantive technical terms (34.2\% selection rate) over common words (14.5\%) and predictable tokens such as numbers (5.5\%); see Appendix~\ref{app:entropy_analysis} for the full distribution analysis and annotated examples.

\noindent\textbf{Valuation via Information Gain.} To evaluate the reasoning quality, we measure the causal contribution of the generated reasoning $z$ to the predictability of these hard tokens. We define the pointwise information gain $g_t(z)$ as the difference in log-likelihood between the reasoning-enhanced policy and the base policy:
\begin{equation}
\resizebox{0.78\hsize}{!}{$
\begin{aligned}
    g_t(z) = & \log \pi_{\theta}(y^*_t \mid x, m, z, y^*_{<t}) \\
             & - \log \pi_{\text{sft}}(y^*_t \mid x, m, y^*_{<t})
\end{aligned}
$}
\end{equation}
The final reward is calculated as the average gain strictly over the selected high-entropy tokens:
\begin{equation}
    \Delta_{IG}(z) = \frac{1}{\sum \mathcal{M}_t} \sum_{t=1}^{|y^*|} \mathcal{M}_t \cdot g_t(z)
\end{equation}
By focusing the gain ($\Delta_{IG}$) exclusively on high-entropy ($\mathcal{M}$) regions, EAIG acts as a microscopic validator. It rewards the model only when its reasoning effectively elucidates the complex technical core of the solution, rather than for memorizing trivial text.

\subsubsection{Contrastive Semantic Gain}

While EAIG ensures the presence of technical details, it does not guarantee that these details form a coherent solution aligned with the research goals. To address this, we introduce Contrastive Semantic Gain (CSG) which acts as a macroscopic guide for the research direction.

\noindent\textbf{Global Intent Alignment.} We first project the texts into a dense semantic space using a pre-trained embedding model $\mathbf{E}(\cdot)$. As illustrated in Figure~\ref{fig:reward} (Right), we calculate the semantic proximity of the generated method $\hat{y}$ to the ground truth $y^*$:
\begin{equation}
    S_{gen} = \text{CosSim}(\mathbf{E}(\hat{y}), \mathbf{E}(y^*))
\end{equation}

\noindent\textbf{Counterfactual Baseline.} A naive similarity check is insufficient because the research context $x$ inherently shares semantic overlap with the solution. To measure the true intellectual increment, we introduce a counterfactual baseline $S_{base}$. This metric represents the semantic score achievable by a trivial policy that merely copies the input:
\begin{equation}
    S_{base} = \text{CosSim}(\mathbf{E}(x \oplus m), \mathbf{E}(y^*))
\end{equation}

\noindent\textbf{Contrastive Valuation.} The final semantic gain is derived from the contrast between the generation and the baseline:
\begin{equation}
    \Delta_{sem} = S_{gen} - S_{base}
\end{equation}
A positive $\Delta_{sem}$ indicates that the model has successfully moved from the problem space ($x \oplus m$) towards the solution space ($y^*$) through its reasoning process. This explicitly rewards the model for making a genuine semantic leap rather than restating background information.

\subsubsection{Reward Synergy}

The core of MoRI lies in the synergy between EAIG and CSG. As depicted in Figure~\ref{fig:reward}, EAIG acts as a microscopic validator ensuring technical depth, while CSG serves as a macroscopic guide for logical direction. This complementary design ensures that the model neither produces fluent but empty platitudes nor hallucinates disconnected terminologies to game the entropy metric.

To filter noise and stabilize optimization, we apply a piecewise step-function shaping $f_{step}(\cdot)$ to both gains (see Appendix~\ref{app:reward_shaping} for details). The final composite reward integrates these shaped signals with two critical regularization terms:
\begin{equation}  
\begin{split}  
R_{total} = \alpha(z) \cdot \mathds{1}[\text{valid}] & \cdot \big(w_e f_{step}(\Delta_{IG}) \\  
& + w_s f_{step}(\Delta_{sem}) \big)  
\end{split}  
\end{equation} 

\noindent\textbf{Length Anchoring $\alpha(z)$.} RL algorithms often exploit short-term rewards by collapsing reasoning chains, reducing variance but compromising depth (see Appendix~\ref{app:length_bias} for details). To counter this cognitive shortcut, we introduce a length-dependent modulation factor:
\begin{equation}
    \alpha(z) = \min \left( 1, 1 - \lambda \frac{L_{anchor} - |z|}{L_{anchor}} \right)
\end{equation}
where $L_{anchor}$ is a pre-defined threshold. This term penalizes the reward if the reasoning length $|z|$ is insufficient to support complex deduction.

\noindent\textbf{Format Constraint $\mathds{1}[\text{valid}]$.} To prevent reward hacking, this indicator function returns 0 if the generated reasoning is empty, shorter than a minimum length threshold, or contains format leakage (\textit{e.g.}, structural headers such as \texttt{\#\#} or \texttt{\#\#\#} that belong to the final output). This strictly enforces the separation between the reasoning process and the final methodology presentation.

\subsection{Inference Procedure}

At inference, MoRI operates in a cascaded manner, as shown in Eq.~\ref{eq:infer}. Given a new research context $x$, the framework first generates a motivation $m$. Then, based on the combined input $x \oplus m$, it performs reasoning $z$ to produce the final methodology $y$. This ensures that the ideation is not a combination of concepts but a structured solution scientifically derived from a clearly defined scientific motivation.

\section{Experiments}
\label{sec:experiments}

In this section, we empirically evaluate the performance of MoRI. Implementation details are provided in Appendix~\ref{app:implementation_details}. Our experiments are designed to address the following research questions:

\begin{itemize}[leftmargin=*, noitemsep, topsep=0pt]
    \item \textbf{RQ1:} Can MoRI generate scientific ideas that are more novel, rigorous, and feasible compared to commercial models and specialized agentic frameworks?
    \item \textbf{RQ2:} How do the proposed composite reward components, specifically the synergy between entropy-aware information gain and semantic alignment, contribute to the optimization of the reasoning process?    
    \item \textbf{RQ3:} Does RL process successfully internalize deep reasoning patterns as intended rather than exhibiting superficial reward hacking behaviors?
\end{itemize}

We first detail the experimental setup followed by the main results to address \textbf{RQ1}. Then we conduct extensive ablation studies and analyze training dynamics to provide insights for \textbf{RQ2} and \textbf{RQ3}.

\subsection{Experimental Setup}

\paragraph{Dataset and Splitting Strategy.}
Derived from ICLR 2024 and 2025 publications, our dataset employs a strict temporal split where 83 ICLR papers published in late 2025 serve as an in-domain test set. To further assess out-of-domain generalization, we additionally collect 67 accepted NeurIPS 2025 papers as a fully held-out OOD test set. The remaining papers yield a total of 8,000 training samples (two per paper, for motivation and method generation respectively), partitioned by paper into two disjoint subsets: 4,000 samples for SFT initialization and 2,000 RL prompts (method generation only) for RL optimization. This separation ensures that the RL stage optimizes on previously unseen contexts. Detailed construction protocols are provided in Appendix~\ref{app:data_construction}.

\paragraph{Baselines.}  
First, we evaluate \textbf{Commercial Models}, including \textit{GPT-4o} and \textit{Claude-3.5-Sonnet}. We instruct them to think before writing to simulate reasoning capabilities \citep{wei2022chain}. Second, we assess \textbf{Agentic Frameworks}, including \textit{AI-Scientist-V2} \citep{yamada2025ai}, \textit{ResearchAgent} \citep{baek2025researchagent}, and \textit{VirSci} \citep{su2025many}. 
To ensure rigorous comparison, we adapted these external baselines to align with our standardized input contexts and task definitions (see Appendix~\ref{app:baseline_adaptation} for details). 
Third, we examine \textbf{Internal Baselines} represented by \textit{Full-SFT}, which operates as a two-stage pipeline ($x \rightarrow m \rightarrow y$) to quantify the specific improvements attributable to the RL optimization. Ideation examples are provided in Appendix~\ref{app:qualitative_examples}.

\begin{table}[t]
\centering
\resizebox{\columnwidth}{!}{%
\begin{tabular}{@{}l cc cc cc cc@{}}
\toprule
\multirow{2}{*}{\textbf{Source}} & \multicolumn{2}{c}{\textbf{Overall}} & \multicolumn{2}{c}{\textbf{Novelty}} & \multicolumn{2}{c}{\textbf{Rigor}} & \multicolumn{2}{c}{\textbf{Feasibility}} \\
\cmidrule(lr){2-3} \cmidrule(lr){4-5} \cmidrule(lr){6-7} \cmidrule(lr){8-9}
& H & L & H & L & H & L & H & L \\
\midrule
MoRI & 2.76 & 3.07 & 2.73 & 3.20 & 2.73 & 3.07 & 2.80 & 2.93 \\
AI-Scientist-V2 & 2.17 & 2.76 & 2.14 & 2.57 & 2.00 & 2.71 & 2.36 & 3.00 \\
GPT-4o & 1.75 & 2.17 & 1.50 & 2.00 & 1.75 & 2.25 & 2.00 & 2.25 \\
Ground Truth & 3.29 & 3.47 & 3.33 & 3.53 & 3.47 & 3.80 & 3.07 & 3.07 \\
\midrule
\rowcolor{gray!12}
\textbf{Overall} & 2.67 & 3.03 & 2.65 & 3.02 & 2.67 & 3.12 & 2.69 & 2.94 \\
\textbf{Pearson $r$} & \multicolumn{2}{c}{0.715} & \multicolumn{2}{c}{0.723} & \multicolumn{2}{c}{0.751} & \multicolumn{2}{c}{0.682} \\
\bottomrule
\end{tabular}%
}
\caption{Human-LLM alignment analysis. We report mean scores from human experts (H) and LLM judge (L) across 60 samples from three dimensions.} 
\label{tab:human_alignment}
\end{table}

\begin{table*}[t]
\centering
\resizebox{\textwidth}{!}{%
\begin{tabular}{@{}l cccc c cccc c cccc c@{}}
\toprule
& \multicolumn{5}{c}{\textbf{ICLR Test Set}} 
& \multicolumn{5}{c}{\textbf{NeurIPS OOD Test Set}} 
& \multicolumn{5}{c}{\textbf{Overall}} \\
\cmidrule(lr){2-6} \cmidrule(lr){7-11} \cmidrule(lr){12-16}
\textbf{Model} 
& Nov. & Rig. & Feas. & Mean & $\Delta$ 
& Nov. & Rig. & Feas. & Mean & $\Delta$ 
& Nov. & Rig. & Feas. & Mean & $\Delta$ \\
\midrule
\rowcolor{gray!8}
\multicolumn{16}{@{}l}{\textit{Commercial Models}} \\
GPT-4o 
& 2.51 & 2.78 & 2.79 & 2.69 & +18.6\% 
& 2.57 & 2.87 & 2.88 & 2.77 & +13.7\% 
& 2.54 & 2.83 & 2.84 & 2.74 & +16.1\% \\
Claude-3.5-Sonnet 
& 3.39 & 3.07 & 2.82 & 3.09 & +3.2\% 
& \textbf{3.42} & \textbf{3.03} & 2.94 & 3.13 & +0.6\% 
& \textbf{3.40} & 3.05 & 2.89 & 3.11 & +2.3\% \\
\midrule
\rowcolor{gray!8}
\multicolumn{16}{@{}l}{\textit{Agentic Frameworks}} \\
AI-Scientist-V2\textsuperscript{\textdagger}(Sonnet) 
& \textbf{3.45} & 3.08 & 2.89 & 3.14 & +1.6\% 
& -- & -- & -- & -- & -- 
& -- & -- & -- & -- & -- \\
AI-Scientist-V2(Haiku) 
& 2.74 & 2.46 & 2.89 & 2.70 & +18.1\% 
& -- & -- & -- & -- & -- 
& -- & -- & -- & -- & -- \\
AI-Scientist-V2(GPT-4o) 
& 2.48 & 2.67 & 2.98 & 2.71 & +17.7\% 
& 2.28 & 2.55 & 2.61 & 2.48 & +27.0\% 
& 2.38 & 2.60 & 2.82 & 2.60 & +22.3\% \\
ResearchAgent 
& 2.66 & 2.51 & 2.63 & 2.60 & +22.7\% 
& 2.58 & 2.22 & 2.32 & 2.37 & +32.9\% 
& 2.63 & 2.39 & 2.50 & 2.50 & +27.2\% \\
VirSci 
& 2.21 & 2.26 & 2.28 & 2.25 & +41.8\% 
& 2.16 & 2.27 & 2.27 & 2.23 & +41.3\% 
& 2.19 & 2.26 & 2.27 & 2.24 & +42.0\% \\
\midrule
\rowcolor{gray!8}
\multicolumn{16}{@{}l}{\textit{Ours}} \\
Full-SFT 
& 3.10 & 2.87 & 3.02 & 2.99 & +6.7\% 
& 2.85 & 2.76 & 2.94 & 2.85 & +10.5\% 
& 2.99 & 2.82 & 2.99 & 2.93 & +8.5\% \\
\textbf{MoRI} 
& 3.31 & \textbf{3.16} & \textbf{3.11} & \textbf{3.19} & -- 
& 3.30 & 3.00 & \textbf{3.16} & \textbf{3.15} & -- 
& 3.31 & \textbf{3.09} & \textbf{3.13} & \textbf{3.18} & -- \\
\midrule
\rowcolor{gray!15}
\textit{Ground Truth} 
& \textit{3.63} & \textit{3.69} & \textit{3.44} & \textit{3.58} & \textit{Ref.} 
& \textit{3.51} & \textit{3.69} & \textit{3.57} & \textit{3.59} & \textit{Ref.} 
& \textit{3.58} & \textit{3.69} & \textit{3.49} & \textit{3.59} & \textit{Ref.} \\
\bottomrule
\end{tabular}%
}
\caption{Main comparison across test sets. We report results on the in-domain ICLR test set (83 papers), the out-of-domain NeurIPS test set (67 papers), and their combination (150 papers). $\Delta$ denotes the relative improvement of MoRI over each baseline. Best results (excluding Ground Truth) are in \textbf{bold}. \textdagger~denotes the use of a stronger backbone (\texttt{claude-3-5-sonnet}) for AI-Scientist-V2. For agentic frameworks, the default backbone is \texttt{gpt-4o-2024-08-06}, with Haiku referring to \texttt{claude-3-5-haiku-2024-10-22}.}
\label{tab:main_results}
\end{table*}

\paragraph{Evaluation Protocol.}  
To address the lack of deterministic verifiers in scientific ideation, we employ a retrieval-augmented LLM judge using \textit{Gemini-2.5-Pro}, grounded in retrieved related works to provide a comprehensive reference for assessing \textit{Novelty}, \textit{Technical Rigor}, and \textit{Feasibility}. To validate reliability, we conducted a human evaluation where three PhD researchers blindly scored 60 samples from diverse sources. The resulting Pearson correlation of 0.715 ($p<0.001$) presented in Table~\ref{tab:human_alignment} confirms that our automated judge reliably approximates expert assessment. Detailed protocols are provided in Appendix~\ref{app:evaluation}. 

\subsection{Performance Comparison}

Table~\ref{tab:main_results} presents the comparative results. MoRI with the optimal configuration ($w_s=0.7, w_e=0.3$, Top-25\% entropy mask) at 400 training steps, achieves the highest mean score of \textbf{3.18}. This constitutes a significant advancement over the Full-SFT baseline (+8.5\%) and outperforms the leading agentic framework.  

On the in-domain ICLR test set, MoRI achieves a mean of 3.19. Notably, while \textit{Claude-3.5-Sonnet} exhibits competitive Novelty (3.39) and \textit{AI-Scientist-V2$^\dagger$(Sonnet)} achieves the highest single-dimension score in Novelty (3.45) by leveraging a stronger backbone, MoRI demonstrates superior capabilities in technical grounding and feasibility. Our model surpasses \textit{Claude-3.5-Sonnet} in Technical Rigor by 2.9\% and achieves a substantial 10.3\% improvement in Feasibility (3.11 vs.\ 2.82). Even against \textit{AI-Scientist-V2$^\dagger$(Sonnet)}, MoRI maintains advantages in Rigor (+2.6\%) and Feasibility (+7.6\%).

On the out-of-domain NeurIPS test set, MoRI maintains a mean of 3.15. While \textit{Claude-3.5-Sonnet} again leads in Novelty, MoRI remains competitive in Technical Rigor and achieves a substantial advantage in Feasibility, resulting in a higher overall mean. This demonstrates the strong generalization of our motivation-grounded reasoning to unseen venues. This observation aligns with recent large-scale evaluations \citep{si2024can, kumar2025can}, confirming that while commercial models can generate conceptually novel proposals, they often lack practical viability.

In contrast, MoRI ensures that generated ideas are both innovative and methodologically sound. The consistent superiority of MoRI over complex workflows such as \textit{AI-Scientist-V2(GPT-4o)} (+22.3\% overall) validates our hypothesis that internalizing the scientific reasoning process via RL yields more robust and generalizable ideation capabilities than external agentic scaffolding.

We further report the combined results with 95\% bootstrap confidence intervals and Bonferroni-corrected pairwise significance tests in Appendix~\ref{app:stat}.

\subsection{Ablation Studies}
\label{subsec:ablation}

To dissect the contribution of each component in MoRI, we conduct controlled ablation studies. We report performance metrics at 100 training steps, a checkpoint where the relative performance trends between configurations stabilize and align with final convergence results, allowing for efficient analysis of optimization dynamics.

\paragraph{Contribution of Motivation Conditioning.}
A key design choice in MoRI is the two-stage motivation-grounded formulation. To isolate its contribution from the RL optimization, we compare three configurations on the NeurIPS OOD test set: (1) Full-SFT with direct generation ($x \rightarrow y$), (2) Full-SFT with two-stage motivation conditioning ($x \rightarrow m \rightarrow y$), and (3) MoRI. As shown in Table~\ref{tab:motivation_ablation}, conditioning on motivation improves over direct generation, validating the formulation itself. MoRI further achieves a significantly larger gain (+0.30 over two-stage SFT), directly isolating the contribution of RL reward design from the motivation-conditioning decomposition.

\begin{table}[t]
\centering
\small
\begin{tabular}{@{}lcccc@{}}
\toprule
\textbf{Model} & \textbf{Nov.} & \textbf{Rig.} & \textbf{Feas.} & \textbf{Mean} \\
\midrule
Full-SFT (Direct) & 2.80 & 2.69 & 2.81 & 2.75 \\
Full-SFT (Two-stage) & 2.85 & 2.76 & 2.94 & 2.85 \\
\textbf{MoRI} & \textbf{3.30} & \textbf{3.00} & \textbf{3.16} & \textbf{3.15} \\
\bottomrule
\end{tabular}
\caption{Ablation on motivation conditioning (NeurIPS OOD test set). Direct: $x \rightarrow y$; Two-stage: $x \rightarrow m \rightarrow y$; MoRI adds RL optimization on top of the two-stage formulation.}
\label{tab:motivation_ablation}
\end{table}

\begin{figure}[t]
    \centering
    \includegraphics[width=1.0\columnwidth]{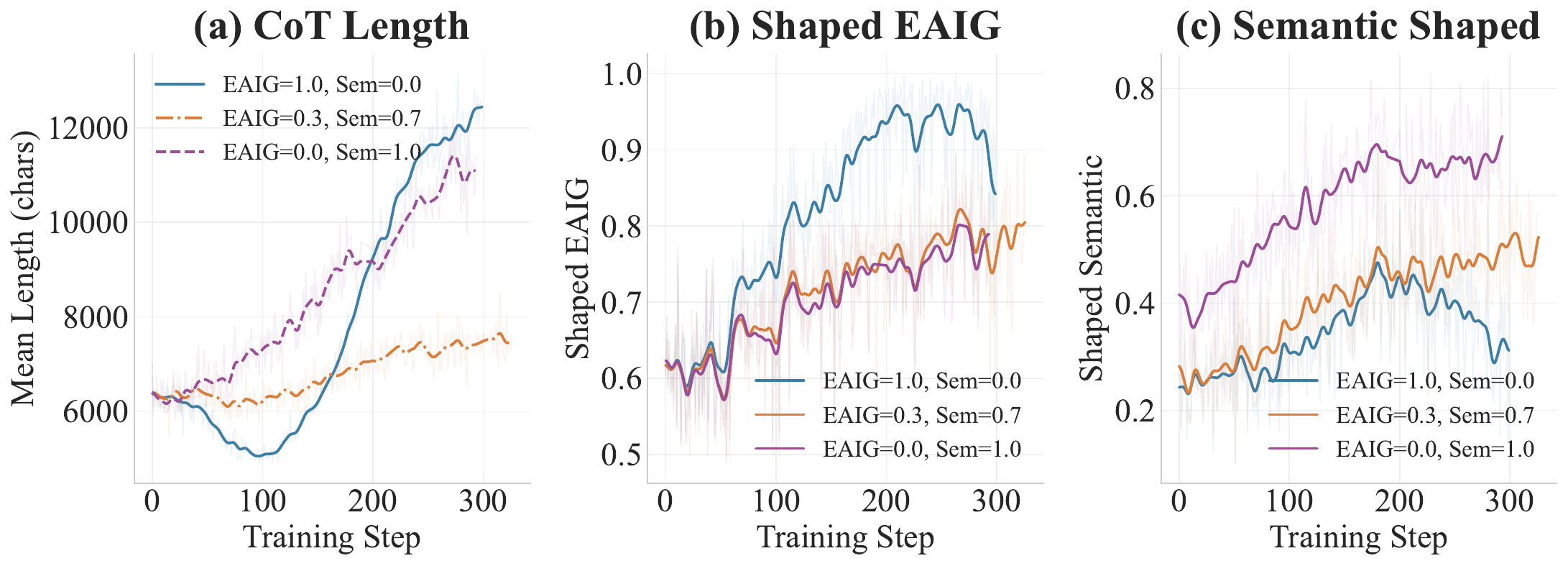}
    \caption{\textbf{Training Dynamics.} Moving average of CoT Length (a), Shaped EAIG (b), and Shaped Semantic Score (c).}
    \label{fig:dynamics}
\end{figure}

\paragraph{Impact of Reward Composition.}
We first investigate the synergy between EAIG and CSG. Table~\ref{tab:ablation_reward} presents the performance metrics, while Figure~\ref{fig:dynamics} visualizes the training dynamics of CoT length and reward values across different configurations.

\begin{table}[ht]
\centering
\resizebox{\columnwidth}{!}{%
\begin{tabular}{@{}cc c cccc@{}}
\toprule
$\boldsymbol{w_s}$ & $\boldsymbol{w_e}$ & \textbf{Description} & \textbf{Novelty} & \textbf{Tech. Rigor} & \textbf{Feasibility} & \textbf{Mean} \\
\midrule
0.0 & 1.0 & EAIG Only & 2.68 & 2.22 & 2.63 & 2.51 \\
1.0 & 0.0 & CSG Only & 3.16 & 2.93 & 3.06 & 3.05 \\
0.5 & 0.5 & Balanced & 3.32 & 2.96 & 2.98 & 3.09 \\
\textbf{0.7} & \textbf{0.3} & \textbf{Optimal} & \textbf{3.34} & \textbf{3.04} & \textbf{3.07} & \textbf{3.15} \\
\bottomrule
\end{tabular}%
}
\caption{Ablation on reward composition (Step 100 w/ Length Anchoring). $w_s$ and $w_e$ denote weights for CSG and EAIG respectively.}
\label{tab:ablation_reward}
\end{table}

As shown in Table~\ref{tab:ablation_reward}, using EAIG in isolation (\texttt{s0-e1}\footnote{We use the notation \texttt{sX-eY} to denote reward weights, where $X$ and $Y$ represent the first decimal digit of $w_s$ (CSG) and $w_e$ (EAIG) respectively.}) leads to severe performance degradation (Mean 2.51). This quantitative failure is explained by the unstable training dynamics shown in Figure~\ref{fig:dynamics} (Blue Line). Although the model initially maximizes EAIG, it eventually suffers from reward collapse and length explosion, where the reasoning trajectory grows uncontrollably and degenerates into incoherent gibberish to hack the entropy metric.

Conversely, the Semantic-only configuration (\texttt{s1-e0}, Purple Line) is more stable but suboptimal. While Figure~\ref{fig:dynamics} (b) and \ref{fig:dynamics} (c) show that optimizing one reward implicitly improves the other, which confirms their synergistic correlation. While the Purple Line still exhibits a rapid rise in reasoning length, indicating a less regulated policy compared to the optimal mix.

The optimal configuration (\texttt{s7-e3}, Orange Line) achieves the best trade-off. It maintains a stable growth in reasoning length and avoids the collapse observed in single-objective settings. This confirms that strong semantic direction combined with a moderate precision constraint provides the necessity for robust scientific reasoning, preventing the model from overfitting to superficial patterns.

\paragraph{Effect of Entropy Mask.}
We further examine the impact of entropy thresholds by comparing the strict top 25\% mask against a looser top 50\% alternative. Table~\ref{tab:ablation_entropy} indicates that tightening the filter yields a 3.7\% overall improvement with Rigor increasing by 0.18 points and Novelty by 0.16 points while maintaining comparable Feasibility.

\begin{figure}[t]
    \centering
    \includegraphics[width=1.0\columnwidth]{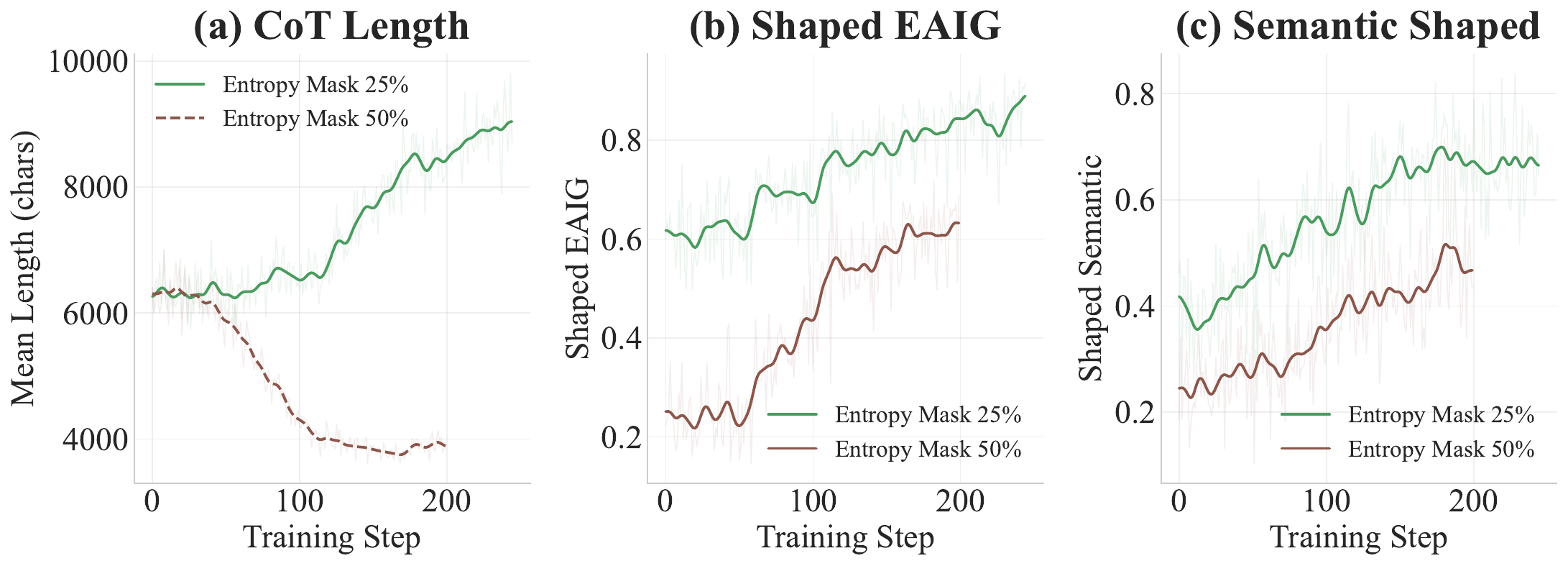}
    \caption{\textbf{Entropy Mask Dynamics.} Moving average of CoT Length (a), Shaped EAIG (b), and Shaped Semantic Score (c).}
    \label{fig:entropy_dynamics}
\end{figure}

\begin{table}[ht]
\centering
\resizebox{\columnwidth}{!}{%
\begin{tabular}{@{}l cccc c@{}}
\toprule
\textbf{Ent. Mask} & \textbf{Novelty} & \textbf{Tech. Rigor} & \textbf{Feasibility} & \textbf{Mean} & $\boldsymbol{\Delta}$ \\
\midrule

Top 50\% & 3.16 & 2.78 & \textbf{3.00} & 2.98 & -- \\
\textbf{Top 25\%} & \textbf{3.32} & \textbf{2.96} & 2.98 & \textbf{3.09} & \textbf{+3.7\%} \\
\bottomrule
\end{tabular}%
}
\caption{Ablation on Entropy Filtering strategies (config $w_s$=0.5, $w_e$=0.5, Step 100).}
\label{tab:ablation_entropy}
\end{table}

The training dynamics in Figure~\ref{fig:entropy_dynamics} elucidate the mechanism. The looser 50\% threshold incorporates functional words that introduce substantial noise into the reward calculation. This interference hinders effective optimization and leads to the reasoning length decline observed in the brown trajectory. Conversely, the top 25\% mask isolates high-complexity technical terms and provides a cleaner signal which sustains the growth of reasoning chains for deeper technical deduction.

\paragraph{Length Anchoring.}
Finally, we analyze the necessity of the Length Anchoring penalty. 

\begin{figure}[t]
    \centering
    \includegraphics[width=1.0\columnwidth]{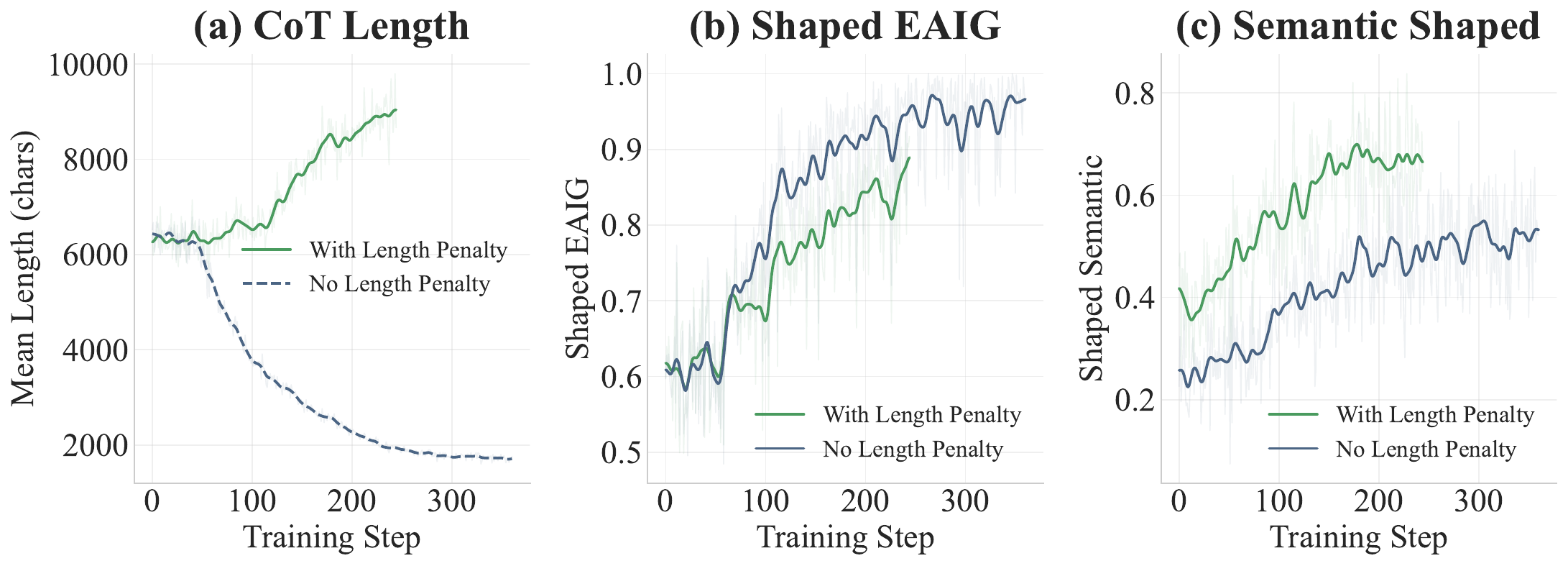} 
    \caption{\textbf{Impact of Length Anchoring.} Moving average of CoT Length (a), Shaped EAIG (b), and Shaped Semantic Score (c).}
    \label{fig:penalty_dynamics}
\end{figure}

\begin{table}[ht]
\centering
\resizebox{\columnwidth}{!}{%
\begin{tabular}{@{}cc cccc c@{}}
\toprule
\textbf{Config} & \textbf{LA} & \textbf{Novelty} & \textbf{Tech. Rigor} & \textbf{Feasibility} & \textbf{Mean} & $\boldsymbol{\Delta}$ \\
\midrule
\multirow{2}{*}{$w_s$=0.7, $w_e$=0.3} & \xmark & 3.22 & 2.94 & 3.00 & 3.05 & -- \\
 & \cmark & \textbf{3.34} & \textbf{3.04} & \textbf{3.07} & \textbf{3.15} & \textbf{+3.2\%} \\
\midrule
\multirow{2}{*}{$w_s$=0.5, $w_e$=0.5} & \xmark & 2.96 & 2.76 & 2.95 & 2.89 & -- \\
 & \cmark & \textbf{3.32} & \textbf{2.96} & 2.98 & \textbf{3.09} & \textbf{+6.9\%} \\
\bottomrule
\end{tabular}%
}
\caption{Impact of Length Anchoring (Step 100). \textbf{LA}: Length Anchoring.}
\label{tab:ablation_length}
\end{table}

Table~\ref{tab:ablation_length} demonstrates that removing Length Anchoring consistently degrades performance across all metrics. In the \texttt{s5-e5} setting, the absence of LA causes a substantial drop in Novelty (3.32 $\to$ 2.96) and Rigor (2.96 $\to$ 2.76). This degradation is elucidated by the training dynamics in Figure~\ref{fig:penalty_dynamics}. Without the penalty, we observe a distinct reasoning collapse where the CoT length drastically decays (Red Line), indicating the model exploits shortcuts to maximize short-term rewards. Conversely, enabling Length Anchoring (Green Line) stabilizes and gradually extends the reasoning process. This enforced depth correlates with higher semantic scores (Figure~\ref{fig:penalty_dynamics} (c)), confirming that the model learns to solve complex ideation tasks through deeper reasoning rather than superficial heuristics. Beyond these quantitative dynamics, we further identify three qualitatively distinct reasoning behavior patterns internalized by MoRI, including goal decomposition, self-critique loops, and paradigm questioning; representative excerpts are provided in Appendix~\ref{app:behavior_patterns}.

\section{Conclusion}
\label{sec:conclusion}

We proposed MoRI, a framework that internalizes scientific ideation via motivation-grounded RL. By modeling the reasoning path from motivations to methodologies, MoRI transcends surface-level imitation. Our composite reward, synergizing entropy-aware information gain with semantic alignment, effectively guides this optimization. Experiments show that MoRI consistently outperforms commercial and agentic baselines in novelty, rigor, and feasibility. Furthermore, analysis of training dynamics confirms that our approach learns robust reasoning patterns rather than shortcuts. This work establishes a promising direction for developing AI systems capable of authentic scientific discovery.

\section*{Limitations}

Our experiments are conducted within computer science, using datasets from machine learning publications. While the motivation-grounded reasoning framework could in principle transfer to other fields, its effectiveness in disciplines with different logical structures, such as biology or physics, remains untested. Evaluation relies on context-aware LLM judges supplemented by human validation on a subset; due to the inherent subjectivity of assessing concepts like scientific novelty, feasibility cannot be fully measured without real-world experiments or large-scale peer review.

While MoRI can generate plausible scientific ideas, it is intended strictly as a collaborative assistant. There is a risk of misuse, such as mass-producing superficial proposals, and the AI cannot evaluate ethical or societal implications. Responsibility for validating the integrity and merit of any generated idea remains with human researchers.

\section*{Ethical Considerations}

MoRI is designed strictly as a collaborative tool to assist human scientists in early-stage ideation, not to replace human creativity or judgment. AI-generated content requires critical evaluation, refinement, and proper attribution, and its use must be transparently disclosed in accordance with guidelines from major venues such as ACL, NeurIPS, and Nature. While AI-assisted ideation can accelerate research, it may also risk homogenizing research directions, underscoring the importance of maintaining human insight and domain expertise. MoRI trains on publicly available papers to internalize reasoning patterns without memorizing or reproducing specific content, ensuring respect for authors’ intellectual contributions and avoiding plagiarism.

We follow a task-driven minimal disclosure policy in constructing our dataset. Although the source papers are publicly available, we remove all author identities, affiliations, citation markers, and reference metadata, retaining only task-relevant background text and abstracted idea-level outputs. This anonymization prevents reverse identification of individual researchers or specific publications while preserving the dataset’s utility for research.

\section*{Acknowledgments}

This work is supported by the National Natural Science Foundation of China (72204087), the Chenguang Program of Shanghai Education Development Foundation and Shanghai Municipal Education Commission (23CGA28), the Shanghai Pujiang Program (23PJC030), Young Elite Scientists Sponsorship Program by CAST (YESS20240562). We also appreciate the constructive comments from the anonymous reviewers.


\bibliography{custom}

\appendix

\begin{figure*}[t]
    \centering
    \includegraphics[width=1.0\textwidth]{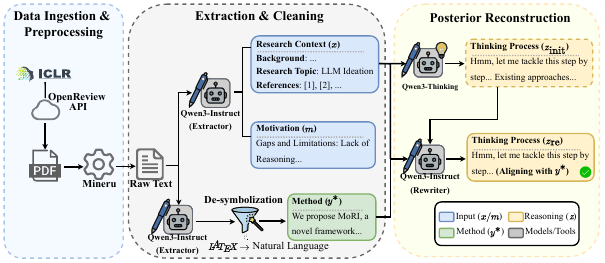} 
    \caption{\textbf{Overview of the Data Construction Pipeline.} The process operates in three distinct phases: (1) \textbf{Data Ingestion \& Preprocessing}, where raw ICLR PDFs are converted into text using MinerU; (2) \textbf{Extraction \& Cleaning}, which parses the research context ($x$) and motivation ($m$), while extracting and \textit{de-symbolizing} the method section ($y^*$) to remove notation-specific noise; and (3) \textbf{Posterior Reconstruction}, where a reasoning chain ($z$) is synthesized by rewriting an initial thought trace ($z_{\text{init}}$) to logically align with the ground-truth method.}
    \label{fig:data}
\end{figure*}

\section{Response Use of LLMs}
In preparing this manuscript, we utilized a large language model exclusively for language polishing and stylistic improvements (\textit{e.g.}, grammar, clarity, and fluency). All scientific content, ideas, experimental design, and results are the original work of the authors. The use of the model did not influence any intellectual contributions, data analysis, or conclusions presented in the paper.

\section{Data Construction and Initialization}
\label{app:data_construction}

We construct a comprehensive data pipeline designed to transform raw scientific papers into structured data for cold start SFT and motivation-grounded RL. As illustrated in Figure~\ref{fig:data}, this pipeline comprises three stages: LLM-based content extraction, reasoning synthesis via posterior reconstruction, and supervised initialization. The resulting dataset $\mathcal{D} = \{(x_i, m_i, z_i, y^*_i)\}_{i=1}^N$, derived from accepted ICLR 2024--2025 papers, serves as the foundation for both the cold start training and the subsequent RL phase.

\paragraph{Data Collection and Extraction.} 
We utilize the OpenReview API\footnote{\href{https://docs.openreview.net/getting-started/using-the-api}{OpenReview API documentation}.} to collect paper PDFs and process them using MinerU \citep{wang2024mineru} to convert papers into Markdown format. To ensure high-quality input, we employ Qwen3-235B-Instruct to parse the raw text, extracting the research topic, background, and main references to form a standardized context $x_i$. 
Simultaneously, we extract the method section as the ground-truth target $y^*_i$. Crucially, to mitigate the stochastic noise introduced by arbitrary variable naming conventions, we perform a \textit{de-symbolization} and cleaning process on $y^*_i$. This involves removing citation markers and rewriting mathematical formulations into natural language descriptions, ensuring the model focuses on the semantic logic of the method rather than overfitting to specific symbolic tokens. The motivation $m_i$ is similarly extracted by the LLM, prompted to identify the specific limitations of prior work addressed by the paper.

\paragraph{Reasoning Synthesis via Posterior Reconstruction.} 
A critical challenge in training scientific agents is the absence of explicit reasoning chains $z$ in papers. We bridge this gap through a two-step \textit{posterior reconstruction} strategy. First, we prompt a strong reasoning model (Qwen3-235B-Thinking) to generate an initial reasoning content based solely on the context $x_i$. Second, to ensure this reasoning logically converges to the ground truth, we employ Qwen3-235B-Instruct to rewrite the initial reasoning content. By conditioning on both the context and the ground-truth method $(x_i, y^*_i)$, the rewriter aligns the reasoning path with the specific proposed solution while maintaining the reasoning style, effectively reverse-engineering the scientific discovery process.

\paragraph{SFT as Cold Start.} 
We initialize our policy model $\pi_{\theta}$ via SFT on two distinct tasks to establish base capabilities: (1) \textbf{Motivation Generation} ($x \to m$), where the model learns to identify research gaps from context; and (2) \textbf{Method Generation} ($x \to (z, y)$), where the model learns to generate the reasoning chain and the method section. Note that in this cold-start stage, the method generation is conditioned only on the context $x$, forcing the model to internalize the dependency between background information and methodological outcomes.

\paragraph{Training Set Splitting and Baselines.}
To strictly evaluate the contribution of RL, we implemented a partitioned training strategy. The total training corpus of 8,000 samples was divided into two disjoint subsets. The first subset containing 4,000 samples was used for supervised initialization described above. The second subset was reserved for RL. Note that while the SFT set includes both motivation and method generation tasks, the RL set specifically targets method generation, resulting in approximately 2,000 valid samples for this stage. This separation ensures that the RL policy explores the reasoning space on contexts it has not encountered during supervised training. Furthermore, the \textbf{Full-SFT} baseline referenced in the main text utilizes the union of these subsets, totaling 8,000 samples, to serve as a high-resource supervised baseline.

\paragraph{Privacy Preservation and Anonymization.}
To mitigate potential privacy and attribution risks associated with redistributing machine-consumable scientific text, we apply a task-driven minimal disclosure policy during dataset construction. Concretely, we remove all personally identifiable and paper-specific metadata, including author names, affiliations, citation markers, and reference entries. The extracted textual inputs only retain task-relevant background content, while the outputs consist of abstracted research ideas at a conceptual level without methodological or experimental details. This anonymization strategy ensures that no individual researcher or original publication can be reverse-identified from the data, while maintaining sufficient information for training and evaluation.

\paragraph{Concrete End-to-End Example.}
Figure~\ref{fig:input_context_example} presents one compact end-to-end example spanning the leakage-controlled input context, the corresponding held-out ground-truth method excerpt, and the inference-time prompt template. This joint presentation makes the data flow explicit: only the context in part (A) is exposed to the model, while the method in part (B) remains hidden supervision and the prompt in part (C) governs method generation.

\begin{figure*}[t]
\centering
\begin{tcolorbox}[
  colback=gray!5,
  colframe=gray!60,
  fonttitle=\bfseries\small,
  title={(A) Leakage-Controlled Input Context},
  boxrule=0.6pt,
  left=6pt,
  right=6pt,
  top=4pt,
  bottom=4pt,
  width=\textwidth,
  fontupper=\footnotesize
]
\textbf{Research Topic.} Improving the ability of Video Large Language Models to recognize and respond appropriately to questions that cannot be answered from the video content.

\medskip
\textbf{Background.} Multimodal Large Language Models have advanced significantly by integrating visual and linguistic data, with Video Large Language Models (Video-LLMs) emerging as a key development. These models combine video understanding with natural language processing to perform tasks such as video-based question answering. They typically rely on large language models as a backbone and are trained to align visual inputs with textual instructions. While progress has been made in enhancing their comprehension of video content, their behavior when faced with questions beyond the scope of the video remains an open issue within the broader field of model alignment and reliability.

\medskip
\textbf{References (title + abstract snippets only).}
\begin{enumerate}[leftmargin=1.5em, noitemsep, topsep=2pt]
    \item \textit{BLIP-2: Bootstrapping Language-Image Pre-training with Frozen Image Encoders and Large Language Models.} The cost of vision-language pre-training has become increasingly prohibitive due to end-to-end training of large-scale models.
    \item \textit{Video-ChatGPT: Towards Detailed Video Understanding via Large Vision and Language Models.} Conversation agents fueled by LLMs provide a new way to interact with video data.
    \item \textit{RLHF-V: Towards Trustworthy MLLMs via Behavior Alignment from Fine-Grained Correctional Human Feedback.} Multimodal LLMs show strong multimodal understanding, reasoning, and interaction capabilities.
    \item \textit{Unanswerable Visual Question Answering.} Recent vision-language models demonstrate strong visual understanding and reasoning, especially on multiple-choice VQA tasks.
    \item \textit{VideoLLaVA: Learning United Visual Representation by Alignment before Projection.} Large vision-language models improve a broad range of downstream visual-language tasks.
\end{enumerate}
\noindent\textit{\ldots}
\end{tcolorbox}

\vspace{3pt}

\begin{minipage}[t]{0.445\textwidth}
\begin{tcolorbox}[
  colback=lightbaseline,
  colframe=baselinecolor,
  fonttitle=\bfseries\small,
  title={(B) Corresponding Ground-Truth Method (Excerpt)},
  boxrule=0.6pt,
  left=6pt,
  right=6pt,
  top=4pt,
  bottom=4pt,
  fontupper=\footnotesize
]
\textbf{Method.} This work presents a framework for improving the ability of Video Large Language Models (Video-LLMs) to recognize when a question cannot be answered based on the visual content of a given video---a capability referred to as \textit{alignment for answerability}. A well-aligned model should not only provide accurate responses to answerable questions but also explicitly identify and decline to answer unanswerable ones, ideally with a valid justification. Without such alignment, models tend to generate confidently incorrect, or hallucinated, answers when presented with questions lacking sufficient visual support.

\medskip
The core challenge lies in the fact that most existing Video-LLMs are trained almost exclusively on datasets where every question is answerable from the video content. As a result, these models are ill-equipped to recognize or respond appropriately to queries that ask about information not present in the video. To address this limitation, the paper proposes a structured approach to align Video-LLMs with the principle of answerability, defines evaluation metrics to assess alignment performance, and introduces a systematic method for constructing training data that includes both answerable and unanswerable question types.

\medskip
\textbf{Preliminaries and Problem Overview.}

\textit{\ldots}

\textbf{Defining Alignment for Answerability.}

\textit{\ldots}
\end{tcolorbox}
\end{minipage}\hfill
\begin{minipage}[t]{0.535\textwidth}
\begin{tcolorbox}[
  colback=gray!5,
  colframe=gray!50,
  fonttitle=\bfseries\small,
  title={(C) Method Generation Prompt (Inference)},
  boxrule=0.5pt,
  left=6pt,
  right=6pt,
  top=4pt,
  bottom=4pt,
  fontupper=\footnotesize
]
You are a world-class AI researcher and scientist with deep expertise in machine learning and a track record of publishing in top-tier conferences.

\medskip
Your primary task is to design a novel and groundbreaking research method that directly addresses the critical research gap and core problem outlined in the Motivation Narrative provided below. Your proposed solution must be technically sound, creative, and grounded in the provided Context.

\medskip
\textbf{Context}

\begin{enumerate}[leftmargin=1.5em, noitemsep, topsep=2pt]
    \item \textbf{Research Topic:} \texttt{\{research\_topic\}}
    \item \textbf{Background:} \texttt{\{background\}}
    \item \textbf{Related Works and References:} \texttt{\{references\}}
\end{enumerate}

\medskip
\textbf{Motivation Narrative (optional for different inference modes)}

This section synthesizes the context, highlights the limitations of existing work, and crystallizes the specific research gap your proposed method must solve.

\texttt{\{motivation\_narrative\}}

\medskip
\textbf{Your Task}

Based on the Context and Motivation Narrative above, write a high-quality research paper draft that proposes an innovative and technically detailed method. Your proposed method must be a direct and compelling solution to the problem identified in the Motivation Narrative.

\medskip
\textbf{Structure your response as follows:}

\textbf{Method}
\begin{itemize}[leftmargin=1.5em, noitemsep, topsep=2pt]
    \item Provide a detailed technical description of your proposed method.
    \item Use clear and precise language and explain the core mechanisms, intuition, and theoretical underpinnings.
    \item Follow the style of top-tier AI academic conferences.
\end{itemize}
\end{tcolorbox}
\end{minipage}

\caption{End-to-end example of one training/inference instance. Part (A) shows the extracted context $x$ provided to the model; part (B) shows an abbreviated excerpt of the corresponding held-out ground-truth method $y^*$; part (C) shows the inference prompt template. Only the context in part (A) is exposed as model input, which prevents leakage of the target method.}
\label{fig:input_context_example}
\end{figure*}

\paragraph{Dataset Scale Justification.}
Our dataset is derived from approximately 4,000 accepted ICLR 2024--2025 papers. Each paper yields two training samples corresponding to the two SFT tasks (motivation generation and method generation), resulting in a total of approximately 8,000 training samples. These are equally partitioned by paper into two disjoint subsets: the first ${\sim}$2,000 papers (${\sim}$4,000 samples) for SFT initialization, and the remaining ${\sim}$2,000 papers for RL optimization. Since RL targets only method generation, this yields approximately 2,000 valid RL prompts, each producing $G{=}16$ rollouts per training step. The training dynamics in Figures~\ref{fig:dynamics}--\ref{fig:penalty_dynamics} further confirm stable convergence well within this budget.

For evaluation, our combined test set of 150 held-out papers yields statistically reliable conclusions: MoRI's 95\% bootstrap confidence interval on the overall score [3.11, 3.25] does not overlap with those of any agentic baseline (upper bounds $\leq 2.70$) or Full-SFT ([2.84, 3.03]), and Bonferroni-corrected pairwise tests confirm significance against all baselines except Claude-3.5-Sonnet (Tables~\ref{tab:ci}--\ref{tab:pairwise}). The ranking of all methods is highly consistent across the two venues, and MoRI's own performance varies minimally between the in-domain ICLR set (3.19) and the out-of-domain NeurIPS set (3.15, $\Delta = 0.04$, n.s.), indicating that 150 papers are sufficient to separate methods reliably. The two test sets also span complementary sub-domain distributions (Table~\ref{tab:subdomain}), ensuring that conclusions are not driven by a narrow set of topics. Scaling to larger corpora and broader scientific disciplines remains an important direction for future work.

\paragraph{Test Set Sub-domain Distribution.}
To characterize the topical coverage of our evaluation, we categorize each test paper into one of six broad ML sub-domains based on its research topic. Table~\ref{tab:subdomain} reports the distribution on both the in-domain ICLR test set and the out-of-domain NeurIPS test set.

\begin{table}[ht]
\centering
\small
\setlength{\tabcolsep}{3pt}

\begin{tabular}{@{}p{4.8cm}cc@{}}
\toprule
\textbf{Sub-domain} & \textbf{ICLR} & \textbf{NeurIPS} \\
\midrule
Perception \& Multimodal & 25.3\% & 16.4\% \\
Language Reasoning \& Alignment & 22.9\% & 19.4\% \\
Generative Modeling & 16.9\% & 7.5\% \\
Learning Efficiency \& Foundations & 16.9\% & 28.4\% \\
Scientific \& Interdisciplinary AI & 10.8\% & 17.9\% \\
Decision, Control \& Safety & 7.2\% & 10.4\% \\
\bottomrule
\end{tabular}
\caption{Sub-domain distribution of the two test sets.}
\label{tab:subdomain}
\end{table}

\section{Reward Shaping Details}
\label{app:reward_shaping}

We apply piecewise step-function shaping to discretize raw gain values into multi-level rewards, which helps filter out noisy low-gain samples while amplifying meaningful improvements. The shaping function is defined as:
\begin{equation}
f_{step}(x; \boldsymbol{\tau}, \mathbf{r}) = \begin{cases}
r_0 & \text{if } x < \tau_1 \\
r_1 & \text{if } \tau_1 \leq x < \tau_2 \\
r_2 & \text{if } \tau_2 \leq x < \tau_3 \\
r_3 & \text{if } x \geq \tau_3
\end{cases}
\end{equation}
where $\boldsymbol{\tau} = [\tau_1, \tau_2, \tau_3]$ are the thresholds and $\mathbf{r} = [r_0, r_1, r_2, r_3]$ are the corresponding reward levels.

Since the information gain $\Delta_{IG}$ and semantic gain $\Delta_{sem}$ operate on different scales, we use separate threshold configurations for each:

\paragraph{Information Gain Shaping.}
For EAIG-based information gain, we use thresholds $\boldsymbol{\tau}_{IG} = [1.0, 1.5, 2.0]$ with reward levels $\mathbf{r} = [0.0, 0.5, 0.8, 1.0]$. These thresholds are calibrated based on the observed distribution of $\Delta_{IG}$ values during training, which typically range from 0 to 2.5 nats.

\paragraph{Semantic Gain Shaping.}
For contrastive semantic gain, we use thresholds $\boldsymbol{\tau}_{sem} = [0.01, 0.05, 0.1]$ with the same reward levels $\mathbf{r} = [0.0, 0.5, 0.8, 1.0]$. These finer-grained thresholds reflect the smaller magnitude of cosine similarity improvements, where a gain of 0.1 already represents substantial semantic enhancement.

\paragraph{Format Validity Constraints.}
The format indicator $\mathds{1}[\text{valid}]$ returns 1 only when all of the following conditions are satisfied:
\begin{itemize}[leftmargin=*, noitemsep, topsep=0pt]
    \item The CoT section is non-empty.
    \item The CoT length exceeds a minimum threshold (1000 characters in our experiments).
    \item The CoT does not contain structured markdown headers (e.g., \texttt{\#\#} or \texttt{\#\#\#}) that indicate format leakage from the method section.
\end{itemize}
These constraints prevent reward hacking behaviors where the model might skip reasoning to directly output method-like content that achieves high semantic similarity with the ground truth.

\subsection{Entropy Distribution and Token Analysis}
\label{app:entropy_analysis}

To validate that the entropy-based mask isolates meaningful scientific content, we analyze the token entropy distribution over 100 training samples (145,521 ground-truth method tokens). The entropy is computed conditioning on the full RL input $(x, m)$ using the fixed SFT model.

\paragraph{Distribution Statistics.}
Table~\ref{tab:entropy_stats} summarizes the entropy distribution. The top-25\% threshold corresponds to an entropy value of 1.93, selecting 36,573 tokens (25.1\%).

\begin{table}[ht]
\centering
\small
\begin{tabular}{@{}lr@{}}
\toprule
\textbf{Statistic} & \textbf{Value} \\
\midrule
Mean $\pm$ Std & 1.26 $\pm$ 1.0 \\
P25 / Median / P75 / P90 & 0.40 / 1.14 / 1.93 / 2.64 \\
EAIG threshold (top-25\%) & 1.93 \\
Selected tokens & 36,573 (25.1\%) \\
\bottomrule
\end{tabular}
\caption{Token entropy distribution statistics over 100 training samples.}
\label{tab:entropy_stats}
\end{table}

\paragraph{Selection Rate by Token Category.}
We classify tokens into six categories using heuristic rules: \textit{technical terms} (alphabetic tokens $\geq$3 characters, e.g., \texttt{attention}, \texttt{gradient}), \textit{common words} (high-frequency function words), \textit{math symbols} (operators and Greek letters), \textit{numbers}, \textit{punctuation}, and \textit{whitespace}. Table~\ref{tab:token_category} reports the selection rate (fraction selected by the top-25\% mask) for each category.

\begin{table}[ht]
\centering
\small
\begin{tabular}{@{}lrcc@{}}
\toprule
\textbf{Category} & \textbf{Count} & \textbf{Sel.\ Rate} & \textbf{Mean Ent.} \\
\midrule
Technical terms & 82,896 & \textbf{34.2\%} & 1.52 \\
Other & 6,307 & 17.2\% & 0.93 \\
Common words & 40,279 & 14.5\% & 0.95 \\
Whitespace & 615 & 12.5\% & 0.86 \\
Math symbols & 2,240 & 10.8\% & 0.73 \\
Punctuation & 12,461 & 7.7\% & 0.87 \\
Numbers & 723 & 5.5\% & 0.61 \\
\bottomrule
\end{tabular}
\caption{Selection rate by token category under the top-25\% entropy mask. Technical terms are selected at more than twice the rate of common words.}
\label{tab:token_category}
\end{table}

Technical terms are selected at 34.2\%, more than twice the rate of common words (14.5\%) and over four times that of numbers (5.5\%). Math symbols and numbers have low selection rates because they follow predictable patterns in scientific writing (e.g., equation structure), while the specific \textit{choice} of technical terms is most likely where genuine scientific novelty resides.

\begin{figure*}[t]
\centering
\includegraphics[width=\textwidth]{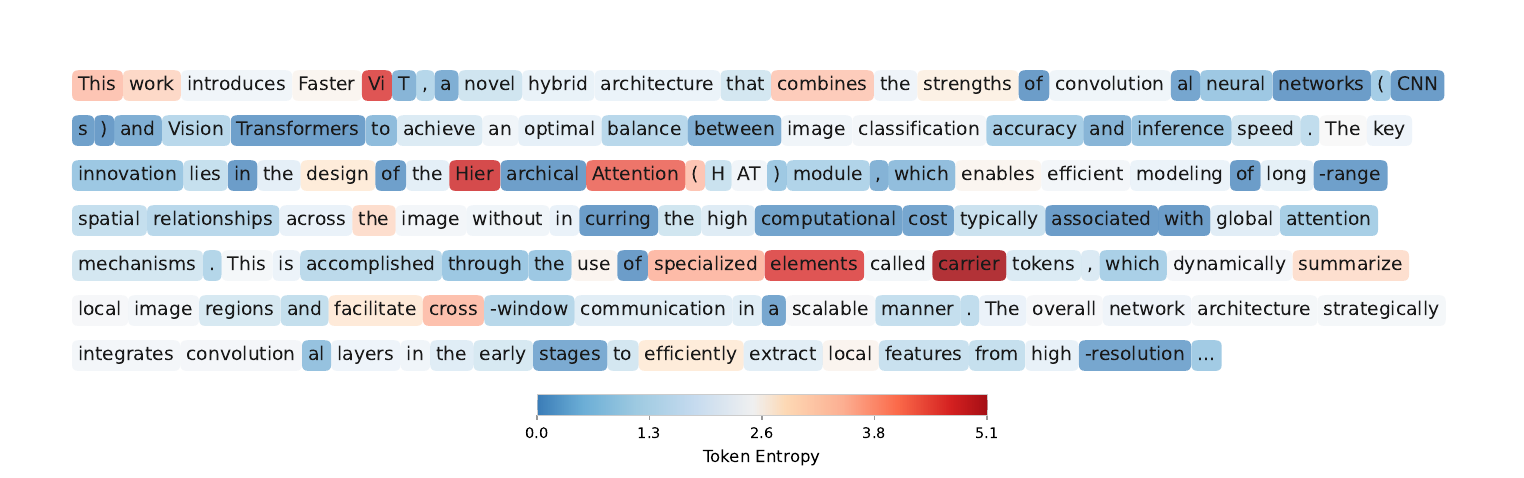}
\caption{Annotated excerpt from a ground-truth method section. Token color indicates entropy (blue=low, red=high).}
\label{fig:entropy_annotated}
\end{figure*}

\paragraph{Annotated Example.}
Figure~\ref{fig:entropy_annotated} presents an annotated excerpt from a ground-truth method section, where each token is labeled with its entropy value and high-entropy tokens (above threshold 1.93) are highlighted. As shown, high-entropy tokens correspond to specific architectural choices and functional design decisions (e.g., ``\textit{carrier tokens}'', ``\textit{Hierarchical Attention}'', ``\textit{dynamically summarize}''), while common function words and predictable structural phrases receive low entropy.

\section{Baseline Adaptation Details}
\label{app:baseline_adaptation}

To ensure a fair comparison, we adapt all baselines to our evaluation setting while preserving their core algorithmic innovations. This section details the modifications made to commercial models and agentic frameworks. All baselines receive identical input contexts and are required to produce detailed Method sections as output.

\subsection{Guiding Principles}

Our adaptations follow three common principles across all baselines:

\noindent\textbf{Input Standardization.} All baselines receive the same research context from our dataset, including the research topic, background information, and related works. This ensures that differences in output quality reflect the capabilities of each method rather than variations in input information.

\noindent\textbf{Output Standardization.} Since our evaluation focuses on the quality of generated methodologies, we require all baselines to produce detailed Method sections of comparable length and structure. For methods that originally output shorter formats such as abstracts or structured proposals, we introduce a post-processing stage to expand the output.

\noindent\textbf{Information Leakage Prevention.} For baselines involving literature retrieval, we implement filtering mechanisms to prevent access to the ground-truth papers. This includes removing paper titles from inputs and filtering search results based on lexical overlap.

\subsection{Commercial Models}

We evaluate GPT-4o and Claude-3.5-Sonnet as representative commercial baselines. To simulate reasoning capabilities comparable to our approach, we design prompts that instruct the model to engage in deliberate thinking before generating the final output.

\noindent\textbf{Prompt Design.} The prompt explicitly separates the reasoning phase from the generation phase. The model is first asked to analyze the research context, identify potential gaps, and outline a solution strategy. Only after this deliberation does the model produce the detailed Method section. This design mirrors the motivation-grounded reasoning process in MoRI.

\noindent\textbf{Generation Prompt for Commercial Models.}
\newtcolorbox{promptbox}[1][]{
  colback=gray!5,
  colframe=gray!50,
  fonttitle=\bfseries\small,
  title=#1,
  boxrule=0.5pt,
  left=6pt,
  right=6pt,
  top=4pt,
  bottom=4pt,
  fontupper=\small,
  breakable
}

\begin{promptbox}[Commercial Models: Generation Prompt]
You are a world-class AI researcher. Given the following research context, your task is to propose an innovative methodology.

\medskip
Before writing the Method section, think deeply about: (1) What are the key limitations in existing approaches? (2) What principles should guide a better solution? (3) What are the core technical components needed?

\medskip
First, write your reasoning process under "\#\# Thinking". Then, write a comprehensive Method section under "\#\# Method" that includes problem formulation, framework overview, key components, and training procedures. The Method section should be 800-1200 words.

\medskip
\textbf{[Research Context]} \texttt{\{context\}}
\end{promptbox}

\noindent\textbf{Parameter Settings.} We use \texttt{gpt-4o-2024-08-06} and \texttt{claude-3-5-sonnet-2024-10-22} with temperature set to 0.7 and maximum output tokens set to 4096.

\subsection{AI-Scientist-V2}

AI-Scientist-V2 \citep{yamada2025ai} employs an agent-based architecture where the model iteratively searches academic literature and refines research proposals through multiple reflection rounds.

\subsubsection{Preserved Core Design}

We retain the following components without modification:

\noindent\textbf{Agent-based Architecture.} The model chooses between two actions at each step: searching Semantic Scholar for relevant papers or finalizing the research proposal. This enables autonomous information gathering.

\noindent\textbf{Iterative Reflection.} The ideation process employs multiple reflection rounds during which the model evaluates and refines its proposals based on novelty and feasibility criteria.

\noindent\textbf{Structured Output.} Generated ideas follow a standardized schema containing hypothesis, related work, abstract, and experimental design.

\subsubsection{Task-Specific Modifications}

\noindent\textbf{Information Leakage Prevention.} We remove paper titles from input contexts and implement a filtering mechanism for Semantic Scholar results. Papers with greater than 80\% lexical overlap with the research topic are excluded. We also restrict searches to papers published before September 2024.

\noindent\textbf{Output Expansion.} The original framework outputs structured proposals rather than detailed Method sections. We add a post-processing stage that expands the proposal into a comprehensive methodology using the prompt shown below.

\noindent\textbf{Method Expansion Prompt.}
\begin{promptbox}[AI-Scientist-V2: Method Expansion Prompt]
Based on the following research idea, write a detailed Method section for a top-tier machine learning conference paper.

\medskip
\textbf{[Research Idea]}\\
Title: \texttt{\{title\}}\\
Hypothesis: \texttt{\{hypothesis\}}\\
Abstract: \texttt{\{abstract\}}

\medskip
Write a comprehensive Method section that clearly describes the proposed approach, includes mathematical formulations where appropriate, explains key components and their interactions, and describes the training and inference procedures. The section should be 800-1500 words.
\end{promptbox}

Table~\ref{tab:ais_params} summarizes the parameter settings.

\begin{table}[ht]
\centering
\small
\begin{tabular}{@{}lll@{}}
\toprule
\textbf{Parameter} & \textbf{Value} & \textbf{Description} \\
\midrule
max\_generations & 1 & Ideas per topic \\
num\_reflections & 3 & Reflection rounds \\
max\_results & 10 & Papers per query \\
year\_filter & -2024 & Publication cutoff \\
overlap\_threshold & 0.8 & Leakage filter \\
\bottomrule
\end{tabular}
\caption{AI-Scientist-V2 parameter settings.}
\label{tab:ais_params}
\end{table}

\subsection{ResearchAgent}

ResearchAgent \citep{baek2025researchagent} generates research ideas through a three-stage pipeline with iterative refinement based on multi-dimensional feedback.

\subsubsection{Preserved Core Design}

\noindent\textbf{Iterative Refinement.} We preserve the generate-validate loop where the Method Developer proposes solutions and the Method Validator provides structured feedback across multiple quality dimensions.

\noindent\textbf{Multi-Dimensional Evaluation.} The validator assesses five dimensions: Clarity, Validity, Rigorousness, Innovativeness, and Generalizability. Each dimension is evaluated independently through parallel processing.

\noindent\textbf{Feedback-Driven Optimization.} When scores fall below the threshold of 5, dimensional feedback is incorporated into refinement prompts to address specific weaknesses.

\noindent\textbf{Best-of-N Selection.} The system selects the highest-scoring method from all iterations as the final output.

\subsubsection{Task-Specific Modifications}

\noindent\textbf{Pipeline Simplification.} Since our dataset provides explicit research topics, we bypass the Problem Identification stage. We also skip the Experiment Design stage to focus evaluation on methodology quality. Only the Method Development stage with its full iterative refinement is executed.

\noindent\textbf{Input Adaptation.} The original system expects Semantic Scholar paper IDs. We implement regex-based extraction to parse research topics, backgrounds, and related works from our dataset format.

Table~\ref{tab:ra_modifications} summarizes the pipeline modifications.

\begin{table}[ht]
\centering
\small
\begin{tabular}{@{}lll@{}}
\toprule
\textbf{Stage} & \textbf{Original} & \textbf{Adapted} \\
\midrule
Problem ID & 3-round iteration & Bypassed \\
Method Dev & 3-round iteration & Preserved \\
Experiment & 3-round iteration & Bypassed \\
Input & API retrieval & Regex parsing \\
\bottomrule
\end{tabular}
\caption{ResearchAgent pipeline modifications.}
\label{tab:ra_modifications}
\end{table}

\subsection{VirSci}

VirSci \citep{su2025many} simulates collaborative research through multi-agent discussions where multiple scientist agents iteratively generate and refine ideas from diverse perspectives.

\subsubsection{Preserved Core Design}

\noindent\textbf{Multi-Agent Discussion.} Multiple agents with distinct academic backgrounds collaborate through iterative discussions, providing complementary viewpoints on research problems.

\noindent\textbf{Novelty Checking.} Generated ideas are compared against a literature database using embedding similarity to identify potential overlaps with existing work.

\noindent\textbf{Literature Retrieval.} The system retrieves relevant papers via FAISS indexing to provide knowledge support for idea generation.

\subsubsection{Task-Specific Modifications}

\noindent\textbf{Topic Injection.} The original system generates research topics through agent discussion. We modify the pipeline to accept predefined topics from our dataset, skipping the topic selection stage.

\noindent\textbf{Output Transformation.} The original output is a short abstract of approximately 200 words. We replace this with a detailed Method section generation stage using GPT-4o.

\noindent\textbf{Model Substitution.} Due to computational constraints, we replace locally deployed LLaMA models with API-accessible alternatives. Discussion agents use GPT-4o-mini, review agents use Qwen-2.5-72B-Instruct, and final generation uses GPT-4o.

Table~\ref{tab:virsci_params} compares the original and adapted configurations.

\begin{table}[ht]
\centering
\small
\begin{tabular}{@{}lll@{}}
\toprule
\textbf{Parameter} & \textbf{Original} & \textbf{Adapted} \\
\midrule
Discussion rounds & 7 & 4 \\
Team members & 3 & 4 \\
Discussion model & LLaMA-8B & GPT-4o-mini \\
Review model & LLaMA-70B & Qwen-72B \\
Final output & Abstract & Method \\
\bottomrule
\end{tabular}
\caption{VirSci configuration comparison.}
\label{tab:virsci_params}
\end{table}

\noindent\textbf{Method Generation Prompt.}
\begin{promptbox}[VirSci: Method Generation Prompt]
Based on the provided Research Topic and Idea from multi-agent discussion, write a comprehensive Method section for this paper.

\medskip
The Method section should be 800-1000 words and cover: (1) Overview of the proposed framework, (2) Problem formulation with mathematical notations if appropriate, (3) Detailed explanation of core components, (4) Training and inference procedures.

\medskip
\textbf{[Research Topic]} \texttt{\{topic\}}

\medskip
\textbf{[Idea from Discussion]} \texttt{\{idea\}}
\end{promptbox}

\section{Evaluation Protocol Details}
\label{app:evaluation}

\subsection{Context-Aware LLM Judge}

We employ \textit{Gemini-2.5-Pro} as the primary automated evaluator. Unlike standard LLM-as-a-Judge approaches \citep{zheng2023judging} that evaluate outputs in isolation, our judge is grounded in a retrieved \textit{Related Work Briefing} to provide domain-specific context for assessment. To ensure stability, we run each evaluation three times and report the averaged scores.

\noindent\textbf{Retrieval Strategy.} For each generated idea, we retrieve relevant papers from Semantic Scholar to construct a reference frame. We implement a hierarchical retrieval strategy that balances recent works, comprising 60\% of the retrieved context, with classic foundational papers accounting for the remaining 40\%. This design ensures the judge can assess novelty against both cutting-edge developments and established baselines.

\noindent\textbf{Evaluation Dimensions.} The judge evaluates three primary dimensions on a 1-5 scale. \textit{Novelty} measures the degree to which the proposed method introduces new concepts compared to existing literature. \textit{Technical Rigor} assesses the soundness and completeness of the technical formulation, including mathematical correctness and algorithmic clarity. \textit{Feasibility} evaluates the practical viability of implementing the proposed method with current resources and techniques.

\noindent\textbf{Evaluation Prompt.} We design a comprehensive evaluation prompt that instructs the LLM judge to adopt a domain expert mindset rather than a simple checklist reviewer. The complete prompt template is shown in Figure~\ref{fig:eval_prompt}. The prompt emphasizes three key principles: judging the intrinsic soundness of ideas based on domain knowledge, assuming competent implementation for omitted standard details, and rewarding elegance while penalizing unnecessary complexity. Additionally, we include a Clarity dimension in the prompt to isolate the influence of writing quality from substantive evaluation. This dimension is excluded from our reported metrics to ensure that scores reflect scientific merit rather than linguistic presentation.

\begin{figure*}[t]
\centering
\begin{tcolorbox}[
  colback=gray!5,
  colframe=gray!60,
  fonttitle=\bfseries\small,
  title=LLM Judge: Context-Aware Evaluation Prompt Template,
  boxrule=0.6pt,
  left=6pt,
  right=6pt,
  top=4pt,
  bottom=4pt,
  width=\textwidth,
  fontupper=\small
]
\textbf{Role and Objective}

As a Senior AI Research and Strategy Analyst, you are tasked with conducting a rigorous evaluation of an AI-generated technical proposal. Your goal is to produce a comprehensive evaluation report, focusing on the proposal's novelty in relation to the provided \textit{Related Work Briefing}, as well as the rigor, soundness, and practical feasibility.

\textbf{Background:} Related Work Briefing: \texttt{\{briefing\_context\}}

\tcbline

\textbf{Part 1: Evaluation Methodology}

\textit{The "Domain Expert" Mindset:} When evaluating, act as a Senior Domain Expert rather than a checklist reviewer.

\begin{itemize}[leftmargin=1.5em, noitemsep, topsep=2pt]
    \item \textbf{Judge the Physics, Not Just the Text:} Base your score on the intrinsic soundness of the idea. If a method is known to be unstable or hard to parallelize, penalize it based on that external knowledge.
    \item \textbf{Fill the Gaps:} If the proposal omits standard implementation details, assume a competent implementation. Do not penalize for brevity unless the missing detail is critical.
    \item \textbf{Evaluate Elegance:} Reward elegance and efficiency. Penalize over-engineering or needlessly complex solutions.
\end{itemize}

\textit{Strategic Feasibility Assessment:} Evaluate from the perspective of a well-funded academic lab. Ask whether the required resources are proportional to the expected gain.

\tcbline

\textbf{Part 2: Likert Scale Rating Rubrics (1-5)}

\textit{Novelty and Originality:}
\begin{itemize}[leftmargin=1.5em, noitemsep, topsep=2pt, label={}]
    \item \textbf{5} Transformative: Introduces a fundamentally new paradigm or solves a long-standing open problem.
    \item \textbf{4} Substantive: A strong advance offering a new perspective. The integration strategy offers a non-obvious solution.
    \item \textbf{3} Iterative: A meaningful but expected improvement. Applies known techniques to a new context.
    \item \textbf{2} Derivative: Naive combination of existing modules without strong justification.
    \item \textbf{1} Redundant: Mere reproduction of known methods.
\end{itemize}

\textit{Technical Rigor:}
\begin{itemize}[leftmargin=1.5em, noitemsep, topsep=2pt, label={}]
    \item \textbf{5} Impeccable: Theoretically airtight. Mathematical modeling accurately represents the problem.
    \item \textbf{4} Solid: Core logic is sound. The theoretical argument holds under expert scrutiny.
    \item \textbf{3} Soft: Plausible but relies on strong, unverified assumptions.
    \item \textbf{2} Flawed: Contains logical fallacies or conflates concepts.
    \item \textbf{1} Broken: Contains fundamental mathematical errors or contradictions.
\end{itemize}

\textit{Feasibility:}
\begin{itemize}[leftmargin=1.5em, noitemsep, topsep=2pt, label={}]
    \item \textbf{5} High Confidence: Aligns with engineering principles. An expert can infer a standard implementation path.
    \item \textbf{4} Non-Trivial: Scientifically sound but complex. Requires advanced engineering skills but achievable.
    \item \textbf{3} Skepticism: Introduces hidden risks (e.g., training instability) that are not addressed.
    \item \textbf{2} Prohibitive: Theoretically valid but practically unwise. Resources disproportionate to gain.
    \item \textbf{1} Impossible: Violates causality or requires non-existent hardware.
\end{itemize}

\textit{Clarity:}
\begin{itemize}[leftmargin=1.5em, noitemsep, topsep=2pt, label={}]
    \item \textbf{5} Publication Ready: Exceptionally clear, well-structured, and precise.
    \item \textbf{4} Clear: Easy to understand. Structure is logical. Minor phrasing issues.
    \item \textbf{3} Passable: Understandable with effort. Some sections may be dense or disorganized.
    \item \textbf{2} Confusing: Poorly structured. Key concepts are undefined or buried.
    \item \textbf{1} Incomprehensible: Incoherent text, severe grammatical errors.
\end{itemize}

\tcbline

\textbf{Part 3: Output Format}

\textit{Comprehensive Evaluation Report:} Provide analysis on (1) Technical Soundness and Rigor, (2) Novelty Analysis synthesizing the briefing with broader knowledge, (3) Practical Feasibility regarding resource proportionality, and (4) Overall Conclusion.

\textit{Structured Scores:} Provide ratings in JSON format:

\begin{lstlisting}
{"evaluation_scores": 
    {
        "novelty_originality": 0,
        "technical_rigor": 0,
        "feasibility_implementability": 0,
        "clarity_quality": 0
    }
}
\end{lstlisting}

\textbf{Task:} Evaluate the following methodology section: \texttt{\{method\_text\}}

\end{tcolorbox}
\caption{Complete prompt template for context-aware LLM evaluation. The prompt instructs the judge to adopt a domain expert mindset and provides detailed rubrics for each evaluation dimension.}
\label{fig:eval_prompt}
\end{figure*}

\subsection{Human Expert Validation}

To verify the reliability of our automated metrics, we conducted a rigorous human evaluation study.

\noindent\textbf{Sample Selection.} We randomly sampled 60 instances from four sources to ensure broad coverage of quality levels: 15 from MoRI, 15 from AI-Scientist-V2, 15 from GPT-4o, and 15 from ground-truth papers. This stratified sampling ensures the evaluation set spans from low-quality to high-quality outputs, which is essential for meaningful correlation analysis.

\noindent\textbf{Annotation Procedure.} Three PhD researchers in machine learning served as expert annotators. Each annotator independently scored all 60 instances on the same three dimensions used by the LLM judge. The evaluation was conducted in a blind manner where annotators did not know the source of each instance. For each sample, we computed the mean score across the three annotators as the final human score.

\noindent\textbf{Alignment Results.} Table~\ref{tab:human_alignment} presents the alignment analysis between human and LLM scores. We observe strong correlations across all dimensions, with Technical Rigor showing the highest alignment ($r=0.751$). The overall Pearson correlation of $r=0.715$ with $p<0.001$ confirms that our LLM-based evaluation reliably captures the quality judgments of human experts. We also note that LLM scores tend to be slightly higher than human scores, which is a known phenomenon in LLM-based evaluation that does not affect the ranking consistency.

\paragraph{Additional OOD Validation on NeurIPS.}
To assess whether the judge remains reliable under venue shift, we additionally conducted human validation on a 60-sample subset from the NeurIPS test set. Table~\ref{tab:human_neurips_alignment} reports the per-dimension Pearson correlations between human consensus and the LLM judge. Although these correlations are slightly lower than those in the mixed-source validation above, they remain consistently positive across all three dimensions, indicating that the judge preserves useful ranking fidelity even on a fully held-out venue.

\begin{table}[ht]
\centering
\small
\begin{tabular}{@{}lc@{}}
\toprule
\textbf{Dimension} & \textbf{Pearson $r$} \\
\midrule
Novelty & 0.613 \\
Technical Rigor & 0.684 \\
Feasibility & 0.629 \\
\bottomrule
\end{tabular}
\caption{Human--LLM alignment on an additional 60-sample NeurIPS subset.}
\label{tab:human_neurips_alignment}
\end{table}

\paragraph{Inter-Annotator Agreement.}
Across the human annotation study, the single-rater reliability is $\mathrm{ICC}(2,1)=0.695$ with 95\% confidence interval $[0.62, 0.76]$, and the within-1 agreement rate is 88.2\%. These values indicate moderate-to-good agreement among annotators and support the stability of the human reference scores used for evaluator validation.

\section{Theoretical Analysis of Length Bias}
\label{app:length_bias}

In this appendix, we provide a theoretical analysis of why GRPO exhibits an implicit bias towards shorter sequences, and how our Length Anchoring mechanism counteracts this effect.

\subsection{GRPO and Implicit Variance Aversion}

GRPO computes advantages through within-group normalization. For a given prompt, the algorithm generates $G$ samples with rewards $r_1, r_2, \ldots, r_G$, and computes the advantage for each sample as:
\begin{equation}
    A_i = \frac{r_i - \bar{r}}{\sigma_r}
\end{equation}
where $\bar{r} = \text{mean}(\{r_j\})$ and $\sigma_r = \text{std}(\{r_j\})$.

Consider two competing strategies: a \textit{long-CoT} strategy $\pi_L$ with reward distribution $r \sim \mathcal{N}(\mu_L, \sigma_L^2)$, and a \textit{short-CoT} strategy $\pi_S$ with $r \sim \mathcal{N}(\mu_S, \sigma_S^2)$. Empirically, longer reasoning chains exhibit higher variance ($\sigma_L > \sigma_S$) because they attempt to cover more technical content, some of which may be uncertain or incorrect.

Under GRPO, high-variance strategies suffer from gradient cancellation. When the reward variance within a group is large, some samples receive strongly positive advantages while others receive strongly negative ones. Over multiple training iterations, these opposing gradients tend to cancel out. Conversely, low-variance strategies produce consistent, moderately positive advantages that accumulate stably across updates.

This mechanism can be understood as an implicit Sharpe Ratio maximization:
\begin{equation}
    \max_\pi \frac{\mathbb{E}[R] - \bar{R}}{\sigma_R} \approx \max_\pi \frac{\mathbb{E}[R]}{\sigma_R}
\end{equation}
The algorithm implicitly favors strategies that achieve high expected reward relative to their variance, which biases learning towards shorter, more conservative outputs.

\subsection{Marginal Returns of Reasoning Length}

Let the CoT length be $L$, and let $k(L)$ denote the number of distinct technical topics covered by reasoning of length $L$, where $k$ increases monotonically with $L$. Each topic $i$ contributes an information gain $X_i$ with expectation $\mu_i$ and variance $\sigma_i^2$.

The expected EAIG reward and its variance can be expressed as:
\begin{equation}
    \mathbb{E}[R] = \frac{1}{N} \sum_{i=1}^{k(L)} n_i \mu_i, \quad \text{Var}[R] = \frac{1}{N^2} \sum_{i=1}^{k(L)} n_i^2 \sigma_i^2
\end{equation}
where $n_i$ is the number of high-entropy tokens associated with topic $i$, and $N$ is the total number of such tokens.

As reasoning length increases, newly covered topics typically have lower expected information gain ($\mu_{k+1} < \bar{\mu}$), leading to diminishing marginal returns:
\begin{equation}
    \frac{\partial^2 \mathbb{E}[R]}{\partial L^2} < 0
\end{equation}
Meanwhile, variance continues to grow approximately linearly. This creates an optimal length $L^*$ beyond which the risk-adjusted return decreases:
\begin{equation}
    \frac{\partial}{\partial L}\left(\mathbb{E}[R] - \beta \cdot \text{Var}[R]\right) \xrightarrow{L > L^*} \text{negative}
\end{equation}

Combined with GRPO's variance aversion, this causes the model to converge to reasoning lengths at or below $L^*$, often shorter than desired.

\subsection{Length Anchoring as a Stabilizer}

To counteract the shortening pressure, we introduce a length-dependent modulation factor:
\begin{equation}
    \alpha(z) = \min \left( 1, 1 - \lambda \frac{L_{anchor} - |z|}{L_{anchor}} \right)
\end{equation}

This creates a positive gradient with respect to length when $|z| < L_{anchor}$:
\begin{equation}
    \frac{\partial R}{\partial L} = \frac{\lambda \cdot R_{base}}{L_{anchor}} > 0 \quad \text{when } |z| < L_{anchor}
\end{equation}

By choosing $\lambda$ appropriately, this gradient can offset GRPO's implicit shortening pressure, stabilizing the reasoning length around the target $L_{anchor}$. In practice, we set $L_{anchor}$ to the average CoT length observed in the SFT model and tune $\lambda$ based on training dynamics.

\subsection{Empirical Validation}

Table~\ref{tab:length_bias_empirical} summarizes the observed CoT lengths under different reward configurations, confirming our theoretical predictions.

\begin{table}[ht]
\centering
\begin{tabular}{@{}lcc@{}}
\toprule
\textbf{Config} & \textbf{EAIG Weight} & \textbf{Final CoT Length} \\
\midrule
\texttt{s0-e1} & 100\% & Shortest (collapses) \\
\texttt{s5-e5} & 50\% & Short \\
\texttt{s7-e3} & 30\% & Moderate \\
\texttt{s1-e0} & 0\% & Preserved \\
\bottomrule
\end{tabular}
\caption{Observed CoT length under different EAIG weights w/o Length Anchoring.}
\label{tab:length_bias_empirical}
\end{table}

Higher EAIG weights amplify the variance effect, accelerating length collapse. Length Anchoring successfully prevents this degradation by providing a stabilizing counter-gradient.

\section{Implementation Details}
\label{app:implementation_details}

We implement MoRI with \texttt{VeRL} \citep{sheng2025hybridflow}.

\subsection{Model Configuration}
We utilize \texttt{DeepSeek-R1-Distilled-Qwen-14B} as the base model for both SFT initialization and RL training. 
During RL, we employ GRPO with a rollout size of $G=16$ samples per prompt. To manage memory efficiency, we use Fully Sharded Data Parallel (FSDP) for the actor and reference models, with parameter and optimizer offloading enabled. The inference engine for rollouts is powered by \texttt{SGLang} with a tensor parallelism size of 1.

\subsection{Hyperparameters}
Table~\ref{tab:hyperparameters} summarizes the key hyperparameters used in our RL training phase.

\begin{table}[ht]
\centering
\resizebox{\columnwidth}{!}{%
\begin{tabular}{@{}lc@{}}
\toprule
\textbf{Parameter} & \textbf{Value} \\
\midrule
\multicolumn{2}{l}{\textit{Training Dynamics}} \\
Global Batch Size & 8 \\
PPO Mini-Batch Size & 4 \\
PPO Micro-Batch Size (per GPU) & 2 \\
PPO Epochs per Rollout & 1 \\
Learning Rate & $5 \times 10^{-7}$ \\
LR Scheduler & Cosine with 10\% warmup \\
KL Coefficient & 0.001 \\
Clip Ratio (Low/High) & 0.2 / 0.28 \\
Max Prompt Length & 5000 \\
Max Response Length & 5000 \\
Loss Aggregation Mode & token-mean \\
\midrule
\multicolumn{2}{l}{\textit{Inference (Rollout)}} \\
Temperature & 1.0 \\
Top-p & 0.95 \\
Number of Rollouts ($G$) & 16 \\
\bottomrule
\end{tabular}%
}
\caption{Hyperparameters for MoRI training.}
\label{tab:hyperparameters}
\end{table}

\subsection{Reward Configuration}
The composite reward function is parameterized as follows:

\begin{itemize}[leftmargin=*, noitemsep, topsep=0pt]
    \item \textbf{Entropy-Aware Information Gain (EAIG):} 
    We set the weight $w_e = 0.3$. The entropy mask threshold is set to the top 25\% to isolate high-complexity tokens. Piecewise shaping is enabled to stabilize optimization.
    
    \item \textbf{Contrastive Semantic Gain (CSG):} 
    We set the weight $w_s = 0.7$. Semantic similarity is computed using \texttt{Qwen3-Embedding-8B}. We calculate the gain contrastively by subtracting the baseline similarity (Prompt-GT) from the generation similarity (Gen-GT). We focus on the method overview section for this calculation to capture macro-level alignment.
    
    \item \textbf{Length Anchoring:} 
    To prevent reasoning collapse, we set the penalty strength $\lambda = 0.5$. The anchor length $L_{anchor}$ is dynamically determined based on the SFT model's average output length.
    
    \item \textbf{Format Constraint:} 
    A strict penalty is applied if the generated reasoning trace contains forbidden structural markers (e.g., \texttt{\#\#}) that should only appear in the final output, ensuring a clean separation between thought and response.
\end{itemize}

\section{Qualitative Examples of Generated Ideas}
\label{app:qualitative_examples}

To provide a comprehensive understanding of the outputs generated by different methods, we present a complete example of a research idea generated by MoRI alongside outputs from two baseline systems (AI-Scientist-V2 and Claude-3.5-Sonnet) given the same research problem on improving LLM reasoning through plan-based guidance.

\paragraph{MoRI Output.} Figures~\ref{fig:mori-motivation}, \ref{fig:mori-reasoning}, and \ref{fig:mori-method-1}--\ref{fig:mori-method-2} showcase the complete output generated by MoRI. Several key observations emerge:

\begin{itemize}[leftmargin=*, nosep]
    \item \textbf{Motivation} (Figure~\ref{fig:mori-motivation}): MoRI produces a well-structured literature review that identifies a clear research gap. The narrative logically progresses from establishing the importance of LLM reasoning, through existing approaches and their limitations, to articulating a specific unaddressed challenge.
    
    \item \textbf{Reasoning} (Figure~\ref{fig:mori-reasoning}): The reasoning trace reveals the model's deliberate problem-solving process. Key insights are highlighted, demonstrating how MoRI systematically analyzes existing methods, identifies opportunities for innovation, and synthesizes ideas from related domains (e.g., meta-learning, reinforcement learning) to formulate a novel approach.
    
    \item \textbf{Method} (Figures~\ref{fig:mori-method-1}--\ref{fig:mori-method-2}): The proposed method is technically detailed and coherent, featuring clearly defined components, training procedures, and optimization strategies. The method directly addresses the gap identified in the motivation.
\end{itemize}

\paragraph{Baseline Comparisons.} For comparison, we present the methods generated by AI-Scientist-V2 (Figure~\ref{fig:ai-scientist-v2-method}) and Claude-3.5-Sonnet (Figure~\ref{fig:claude-sonnet-method}) for the same research problem.

\begin{itemize}[leftmargin=*, nosep]
    \item \textbf{AI-Scientist-V2} (Figure~\ref{fig:ai-scientist-v2-method}): The output attempts to integrate numerous trending concepts (schema generation, zero-shot CoT, PPO, DPO, etc.) into a unified framework. However, this aggregation-heavy approach lacks a coherent core insight, with some technical components being redundant or even conflicting (e.g., PPO and DPO). The writing also contains vague terminology and unclear mechanisms that obscure the actual contribution.
    
    \item \textbf{Claude-3.5-Sonnet} (Figure~\ref{fig:claude-sonnet-method}): The method demonstrates polished technical writing with formal notation and structured formulation. However, it presents incompatible implementation paths (prompt-based vs. architecture modification) without resolution.
\end{itemize}

\paragraph{Summary.} These examples illustrate the distinctive characteristics of MoRI: (1) generating research ideas grounded in systematic reasoning rather than direct generation, (2) producing technically detailed and implementable methods, and (3) maintaining coherence between motivation identification and proposed solutions through explicit reasoning traces.

\section{Statistical Significance Analysis}
\label{app:stat}
To verify the statistical reliability of our main results, we conduct two complementary analyses on the combined test set (ICLR + NeurIPS).

\paragraph{Bootstrap Confidence Intervals.}
We compute 95\% bootstrap confidence intervals via 10,000 resamples. As shown in Table~\ref{tab:ci}, MoRI achieves an overall score of 3.18 [3.11, 3.25], whose lower bound exceeds the upper bound of all agentic baselines and the point estimate of Full-SFT (2.93). Notably, MoRI's confidence interval does not overlap with those of any agentic framework, confirming that the improvements are statistically robust.

\begin{table}[ht]
\centering
\resizebox{\columnwidth}{!}{%
\begin{tabular}{@{}l cccc@{}}
\toprule
\textbf{Model} & \textbf{Novelty} & \textbf{Tech.\ Rigor} & \textbf{Feasibility} & \textbf{Overall} \\
\midrule
GPT-4o & 2.54 [2.38, 2.70] & 2.83 [2.71, 2.94] & 2.84 [2.72, 2.97] & 2.74 [2.64, 2.84] \\
Claude-3.5-Sonnet & 3.40 [3.28, 3.52] & 3.05 [2.92, 3.18] & 2.89 [2.78, 2.99] & 3.11 [3.02, 3.20] \\
\midrule
AIS-v2 (GPT-4o) & 2.38 [2.23, 2.53] & 2.60 [2.48, 2.71] & 2.82 [2.69, 2.95] & 2.60 [2.49, 2.70] \\
ResearchAgent & 2.63 [2.46, 2.80] & 2.39 [2.22, 2.56] & 2.50 [2.34, 2.67] & 2.50 [2.35, 2.66] \\
VirSci & 2.19 [2.08, 2.30] & 2.26 [2.16, 2.36] & 2.27 [2.18, 2.37] & 2.24 [2.16, 2.32] \\
\midrule
Full-SFT & 2.99 [2.85, 3.13] & 2.82 [2.71, 2.93] & 2.99 [2.87, 3.11] & 2.93 [2.84, 3.03] \\
\textbf{MoRI} & \textbf{3.31} [3.19, 3.42] & \textbf{3.09} [2.99, 3.19] & \textbf{3.13} [3.02, 3.25] & \textbf{3.18} [3.11, 3.25] \\
\midrule
\rowcolor{gray!15}
\textit{Ground Truth} & \textit{3.58} [3.48, 3.68] & \textit{3.69} [3.59, 3.80] & \textit{3.49} [3.35, 3.64] & \textit{3.59} [3.51, 3.67] \\
\bottomrule
\end{tabular}%
}
\caption{Combined results (ICLR + NeurIPS, 150 papers) with 95\% bootstrap confidence intervals (10,000 resamples). Best results (excluding Ground Truth) are in \textbf{bold}.}
\label{tab:ci}
\end{table}

\paragraph{Pairwise Significance Tests.}
We conduct Bonferroni-corrected Welch's $t$-tests between MoRI and each baseline on the combined overall scores. As shown in Table~\ref{tab:pairwise}, MoRI consistently outperforms most of the agentic frameworks and commercial models ($p < 0.001$, Cohen's $d > 0.8$) and Full-SFT ($p = 0.003$, $d = 0.458$). The comparison with Claude-3.5-Sonnet is not significant ($p =$ n.s.), consistent with our finding that their advantages are complementary: Sonnet excels in Novelty while MoRI leads in Feasibility.

\begin{table}[t]
\centering
\small
\begin{tabular}{@{}lrcc@{}}
\toprule
\textbf{MoRI vs} & $\boldsymbol{\Delta}$ & $\boldsymbol{p}$ & \textbf{Cohen's} $\boldsymbol{d}$ \\
\midrule
VirSci & +0.93 & $<$0.001 & 2.000 (large) \\
ResearchAgent & +0.67 & $<$0.001 & 0.939 (large) \\
AIS-v2 (GPT-4o) & +0.58 & $<$0.001 & 1.083 (large) \\
GPT-4o & +0.44 & $<$0.001 & 0.874 (large) \\
Full-SFT & +0.24 & 0.003 & 0.458 (small) \\
Claude-3.5-Sonnet & +0.07 & n.s. & --- \\
\bottomrule
\end{tabular}
\caption{Pairwise significance tests on the combined set (150 papers). $\Delta$: difference in overall mean. $p$-values are Bonferroni-corrected.}
\label{tab:pairwise}
\end{table}

\begin{figure*}[t]  
\begin{motivationbox}  
\begin{tikzpicture}[remember picture, overlay]
\node[rotate=45, scale=16, text opacity=0.3, gray!70, font=\bfseries\sffamily] 
    at ([yshift=0cm]current page.center) {MoRI};
\end{tikzpicture}
The advent of large language models (LLMs) has catalyzed a paradigm shift in artificial intelligence, enabling remarkable advances across diverse domains such as text understanding, code generation, and even mathematical reasoning. Despite their broad capabilities, the core challenge of \textbf{complex multi-step reasoning} remains a significant bottleneck, particularly in specialized domains like mathematics. A pivotal insight from foundational work has established that LLMs possess an implicit capacity for reasoning, as demonstrated by the success of zero-shot Chain-of-Thought (CoT) prompting, which allows models to generate step-by-step solutions without explicit fine-tuning [Paper 1]. This breakthrough suggests that reasoning ability can be unlocked and improved not through architectural modification, but through strategic prompting strategies that guide the model's internal cognitive process.  
  
In response to this potential, a growing body of research has explored techniques to enhance LLM reasoning through self- or semi-supervised methods. Paper 2 demonstrates that language models can be trained to produce high-quality reasoning paths by using self-generated step-by-step solutions and answer feedback in a bootstrapping framework. Similarly, Paper 3 leverages rejection sampling to generate more reliable reasoning paths by sampling from the model and filtering out low-quality outputs, effectively aligning the model's behavior with desired reasoning qualities. These methods have shown significant improvements in reasoning benchmarks, particularly in Paper 4, which evaluates mathematical problem-solving skills. While these approaches represent important strides in improving LLM reasoning, they face a critical limitation: the generation of reasoning paths often suffers from \textbf{false-positive solutions}---erroneous step-by-step plans that the model eventually corrects. Such inaccuracies introduce inefficiency and inconsistency into the training process, undermining the reliability of the learned reasoning strategies.  
  
This limitation points to a deeper issue: the current prompting paradigms do not explicitly guide the model in \textit{how} to approach a problem before generating the solution. In contrast, human problem-solving behavior is characterized by a deliberate \textit{planning} stage---introducing meta-cognitive abstraction to anticipate and structure the solution process [Paper 5]. For instance, when faced with a complex mathematical problem, a human might first outline a series of subtasks or high-level reasoning steps, and then execute them systematically. By reducing cognitive load through anticipatory abstraction, humans are able to allocate resources more effectively and avoid the inefficiencies of trial-and-error reasoning. This observation suggests a promising direction for improving LLM reasoning: introducing a \textit{planning} phase in which the model generates an abstract, high-level strategy before committing to the detailed execution of a solution.  
  
Although several prompting strategies have attempted to incorporate high-level guidance, they do so implicitly or without explicit structure. For example, Paper 6 separates planning and execution into two distinct phases, and Paper 7 leverages hints to guide reasoning, while Paper 8 uses progressively refined guidance to refine model responses. However, these methods rely on external hints or manually defined prompts, and do not enable the model to \textit{learn} or \textit{generalize} such strategies from its own experience. This stands in contrast to the ultimate goal of \textbf{autonomous self-improvement}, where the model can discover and internalize efficient reasoning plans through its own interactions and feedback.  
  
A key theoretical foundation supporting this direction lies in the principles of meta-learning, particularly Paper 9, which enables models to learn how to learn by acquiring abstract, transferable knowledge that applies across multiple tasks. This concept aligns with the idea of learning \textit{plans}---meta-knowledge that can guide the solution of diverse reasoning problems---rather than task-specific solutions. Furthermore, Paper 10 demonstrates that learning generalizable strategies across tasks can be achieved through efficient adaptation, providing a methodology for extracting and generalizing abstract plans.  
  
Thus, a fundamental gap remains: \textbf{there is no existing method that explicitly extracts, generates, and trains on high-quality, abstract plans that guide the reasoning process of an LLM.} While current approaches improve the quality of step-by-step solutions, they fail to exploit the potential of plan-based guidance to reduce cognitive load, avoid false-positive solutions, and enable more efficient, structured reasoning. Addressing this gap requires a new paradigm---one that moves beyond generating sequences of tokens or fixed-length reasoning steps, and instead focuses on learning and optimizing \textit{anticipatory plans} as a form of meta-knowledge. Such an approach could unlock significant improvements in LLM reasoning by emulating the human ability to plan before acting, thereby enabling more efficient, accurate, and self-improving reasoning systems.  
\end{motivationbox}  
\caption{Generated \textbf{Motivation} by MoRI for the example research idea on plan-based reasoning improvement.}  
\label{fig:mori-motivation}  
\end{figure*}

\begin{figure*}[t]  
\begin{reasoningbox}  
\begin{tikzpicture}[remember picture, overlay]
\node[rotate=45, scale=15, text opacity=0.3, gray!70, font=\bfseries\sffamily] 
    at ([yshift=0cm]current page.center) {MoRI};
\end{tikzpicture}
Okay, I'm faced with this complex problem of improving large language models (LLMs) through plan-based guidance for better reasoning. The goal is to design a novel method that directly addresses the critical gap highlighted in the Motivation Narrative: the lack of a method that extracts, generates, and trains on high-quality, abstract plans that guide LLM reasoning. [\ldots]  
  
\insight{I need to propose a method that shifts the paradigm from generating detailed step-by-step solutions to \textit{learning and optimizing abstract plans}.} This means my approach should go beyond simply improving the quality of solutions; it should fundamentally alter the model's reasoning strategy by first teaching it to \textit{anticipate} and \textit{structure} its solution process. [\ldots]  
  
Let me start by dissecting the state-of-the-art to identify where my novel idea can fit. Paper 2 achieves self-training by generating step-by-step rationales and using answer feedback to improve future generations. [\ldots] Both methods, however, operate at the level of step-by-step execution and do not incorporate any form of high-level planning. [\ldots]  
  
\insight{My proposed method, LEPA (Learning via Episodic Plan Abstraction), must bridge this gap. It should be a data generation and training strategy that integrates plan-based guidance into the self-training loop.} The key innovation is to decouple the generation of a plan from the solution execution. That is, the model first generates a high-level plan in response to the problem, and then solves the problem by following that plan. [\ldots]  
  
I recall that in reinforcement learning, a common goal is to learn a policy that maximizes cumulative reward over time. Similarly, in meta-learning, the objective is to learn a set of parameters that can quickly adapt to new tasks with limited data. \insight{Drawing inspiration from these paradigms, I can conceptualize the plan as a meta-policy---the model's strategy for solving \textit{how} to solve a given problem.} By training on a large corpus of such meta-policies, the model learns to generate high-quality plans that guide subsequent reasoning. [\ldots]  
  
Now, how do I implement this? \insight{I need a two-phase process: first, generating a corpus of high-quality plans using the current LLM, and second, training a new LLM on this corpus to produce better plans.} This forms an iterative feedback loop: better planning leads to better solutions, which in turn leads to better plans. [\ldots]  
  
Once the plan generator is trained, I replace the original LLM with this improved model and restart the self-training loop. Now, instead of generating full solutions, the new model first generates a plan in response to the problem, and then uses that plan to generate a step-by-step solution. \insight{This two-phase process---plan first, solve second---mirrors the human cognitive process} and enables the model to leverage its newly acquired meta-knowledge for more structured reasoning. [\ldots]  
  
A critical design decision is whether to use supervised fine-tuning or a more modern optimization method like Direct Preference Optimization (DPO) for training the plan generator. DPO is a lightweight alternative to reinforcement learning that aligns model outputs with human preferences without requiring complex reward modeling or sampling. [\ldots]  
  
In summary, \insight{LEPA represents a fundamental shift in LLM training---from optimizing token-level execution to optimizing meta-level planning.} By decoupling plan generation from solution execution and treating the plan as a form of learned meta-knowledge, LEPA enables the model to emulate human-like anticipatory reasoning.  
\end{reasoningbox}  
\caption{Generated \textbf{Reasoning} by MoRI (representative excerpts with key insights highlighted).}  
\label{fig:mori-reasoning}  
\end{figure*}

\begin{figure*}[t]  
\begin{methodbox}  
\begin{tikzpicture}[remember picture, overlay]
\node[rotate=45, scale=15, text opacity=0.3, gray!70, font=\bfseries\sffamily] 
    at ([yshift=0cm]current page.center) {MoRI};
\end{tikzpicture}
\textbf{\S~Method}   
  
This paper introduces LEPA, a novel framework for improving large language models (LLMs) through plan-based guidance for multi-step reasoning tasks. LEPA shifts the paradigm from directly optimizing the generation of step-by-step solutions to explicitly learning and refining high-level plans that guide the model's reasoning process. By treating plans as a form of meta-knowledge—abstract, transferable strategies for solving problems—LEPA enables the model to emulate the human cognitive strategy of anticipating and structuring solutions before execution. This approach not only enhances the accuracy of generated solutions but also reduces computational overhead by aligning the planning phase with the solution phase, thereby minimizing the need for extensive re-computation.  
  
\textbf{\S~Overview and Core Idea.}   
  
LEPA operates in an iterative self-training loop, where the model continuously improves its reasoning capabilities by generating and refining high-quality plans. The process is divided into two distinct stages: plan generation and plan-based solution generation. In the first stage, a preliminary plan is generated for a given problem in response to a prompt that asks the model to describe the overall approach to solving the task. This plan is then used in the second stage to guide the generation of a detailed, step-by-step solution. The key innovation is that the plan generator—responsible for producing these high-level strategies—is itself trained using high-fidelity solutions generated in previous iterations. This creates a feedback loop where the quality of the plans improves over time, leading to better reasoning performance. By learning to anticipate the optimal solution path, the model is able to avoid inefficient trial-and-error reasoning and produce more accurate and coherent solutions.  
  
\textbf{Decoupling Plan Generation and Solution Execution.}   
  
A central component of LEPA is the decoupling of plan generation from the generation of detailed solutions. Instead of directly prompting the model to solve a problem, LEPA first asks the model to generate a high-level plan that outlines the main steps required to arrive at the correct answer. This plan is typically structured as a concise list of key steps or subtasks, and it represents a simplified, abstract representation of the reasoning process. The model is prompted to describe how it would approach the problem, not how it would execute the solution. This distinction allows the model to focus on the strategic aspect of problem-solving—anticipating the necessary actions—rather than on the detailed mechanics of execution. As a result, the plan acts as a cognitive roadmap that guides the subsequent solution generation.  
  
The separation of these two phases offers several advantages. First, it enables the model to exploit its ability to generate high-level summaries, which may be more accurate than step-by-step details due to reduced cognitive load. Second, it allows for the explicit evaluation and refinement of plans, which can be used to improve future plan generation. Third, it facilitates the development of a dedicated plan generator that is specialized in producing structured, high-fidelity plans, rather than one that is constrained by the limitations of executing solutions in a single pass.  
  
\textbf{\S~Training the Plan Generator.}   
  
The plan generator is a model that is trained to produce high-quality plans based on the target solution to a given problem. This training is conducted independently of the main self-training loop, using a large corpus of problems and their corresponding step-by-step solutions. The objective is to teach the plan generator to accurately predict the underlying structure of the solution process for any given problem.  
  
During training, each problem is presented to the plan generator along with its correct, high-fidelity solution. The generator is then prompted to produce a plan that reflects the main reasoning steps required to solve the problem. The training data consists of pairs: one for each problem, containing the problem statement and the corresponding plan. The model is trained to minimize the discrepancy between the generated plan and the target solution, using a standard supervised fine-tuning objective. Specifically, the loss function measures how closely the generated plan matches the actual solution in terms of content, structure, and logical flow.  
  
To ensure the quality of the training data, a filtering step is applied to the initial set of solutions. Solutions that are incomplete, contain logical errors, or fail automated validation checks are discarded. Only those solutions that are fully correct and coherent are retained for training. This ensures that the plan generator learns from reliable examples and produces plans that are both accurate and meaningful. The resulting model—referred to as the plan generator—is then integrated into the main self-training loop, where it generates plans that guide the solution generation process.  
\end{methodbox}  
\caption{Generated \textbf{Method} by MoRI (Part 1: Overview, Core Idea, and Training).}  
\label{fig:mori-method-1}  
\end{figure*}

\begin{figure*}[t]  
\begin{methodbox}  
\begin{tikzpicture}[remember picture, overlay]
\node[rotate=45, scale=15, text opacity=0.3, gray!70, font=\bfseries\sffamily] 
    at ([yshift=0cm]current page.center) {MoRI};
\end{tikzpicture}
\textbf{\S~Plan-Based Solution Generation.}   
  
In the solution generation phase, the plan generator is used to produce a high-level plan for a given problem. This plan is then used to guide the generation of a step-by-step solution by structuring the reasoning process according to the abstract steps outlined in the plan. The solution is generated in a two-phase procedure: first, a plan is generated, and second, the model solves the problem by executing that plan.  
  
The solution generation process is executed in two steps. In the first step, the model receives the problem statement and a prompt that asks it to generate a plan. This prompt is designed to elicit a concise, structured description of the overall reasoning strategy. The model responds with a sequence of tokens that represent the plan. These tokens are then processed to extract the relevant plan content—such as a list of key steps or subtasks—and this content is stored as the plan.  
  
In the second step, the model generates the step-by-step solution. To ensure consistency and accuracy, the plan is reused across multiple solution generations. That is, once a plan is generated, it is applied to solve the same problem multiple times, with the solution generated in each case based on the same plan. This allows for multiple evaluations of the plan's effectiveness without requiring it to be regenerated each time. The solution is generated using the original problem statement and the extracted plan, which provides a structured context for the reasoning process.  
  
After generating a solution, the model evaluates its correctness. If the solution is correct, the plan used to generate it is considered high-quality and is retained for future training. If the solution is incorrect, the plan is deemed poor and is discarded. This evaluation step forms a feedback mechanism that improves the quality of future plans by selectively retaining those that lead to accurate solutions. Over time, this iterative process results in the plan generator learning to produce increasingly accurate and effective plans, leading to improved reasoning performance across a wide range of tasks.  
  
\textbf{\S~Evaluating Plan Quality.}   
  
To assess the quality of plans, LEPA introduces a method for evaluating how well a generated plan corresponds to the target solution. This is done by comparing the structure and content of the generated plan to the actual steps in the solution. A similarity score is computed that quantifies the degree of alignment between the two. Plans that closely match the solution in terms of the order and nature of the steps are scored higher, indicating higher fidelity.  
  
The similarity between the plan and the solution is measured using a probabilistic approach based on language modeling. Specifically, a reference model—trained to maximize the probability of sequences of tokens given the problem context—is used to estimate the likelihood of the generated plan given the solution. The higher the probability, the more similar the plan is to the solution, and the higher the similarity score assigned to it. This score reflects how well the plan captures the essential reasoning steps required to solve the problem.  
  
This evaluation method allows for a principled comparison of different plans and provides a quantitative measure of plan quality. It supports the refinement of the plan generator by rewarding high-fidelity plans and penalizing those that deviate significantly from the correct solution. By integrating this evaluation into the training process, LEPA ensures that only the most accurate and coherent plans are used to guide future solution generation, thereby enhancing the overall effectiveness of the self-training loop.  
  
\textbf{Training the Plan Generator Using DPO.}   
  
To further enhance the training of the plan generator, LEPA supports the use of Direct Preference Optimization (DPO), a modern optimization method that aligns model outputs with human preferences. DPO enables the model to learn from human judgments about the quality of plans without requiring explicit human feedback at inference time. Instead, preferences are derived from the internal behavior of the model itself, making the training process more efficient and scalable.  
  
In this approach, the plan generator is trained to produce plans that maximize the likelihood of generating correct solutions, as inferred by a reference model. The training objective is structured such that the model learns to prefer plans that are more likely to lead to accurate solutions. This is achieved by aligning the plan generator's parameters with the preferences of a second, reference model that is used to evaluate solution correctness.  
  
During training, the plan generator is presented with a problem and its correct solution. It is then prompted to generate a plan, and the likelihood of that plan being correct is estimated by the reference model based on its evaluation of the solution. The training loss is computed to encourage the plan generator to produce plans that are more likely to result in correct solutions, according to the reference model. The optimization process adjusts the plan generator's parameters to minimize this loss, effectively training it to generate plans that are more aligned with the knowledge and preferences captured by the reference model.  
  
This method leverages the internal representations of a well-performing model to guide the training of the plan generator, avoiding the need for explicit human annotations. It also supports the use of a larger reference model than the one used to generate solutions, which can provide more accurate evaluations. By incorporating DPO into the training process, LEPA can achieve faster convergence and higher-quality plans, further improving the overall performance of the reasoning system.  
\end{methodbox}  
\caption{Generated \textbf{Method} by MoRI (Part 2: Solution Generation, Evaluation, and DPO Training).}  
\label{fig:mori-method-2}  
\end{figure*}

\begin{figure*}[t]
\begin{baselinebox}{Method (AI-Scientist-V2)}
\textbf{High-Level Planning and Contextual Execution (HLPCE) Framework}

\textbf{1. Step-wise Plan Schema Generation:}
\begin{itemize}[leftmargin=*, nosep]
    \item Develop a two-tiered model structure, starting with an abstract planning schema generation followed by detailed solution pathways.
    \item Use cognitive models and meta-learning techniques to extract planning schemas from broad datasets, offering versatile templates for large language models (LLMs) to base future context-specific implementations.
    \item Introduce specific procedures for the creation and application of these schemas, such as using clustering techniques for organizing schemas based on domain-specific structural differences.
\end{itemize}

\textbf{2. Enhanced Zero-Shot Reasoning \& Prompt Guidance System:}
\begin{itemize}[leftmargin=*, nosep]
    \item Incorporate a sophisticated hint augmentation mechanism inspired by methodologies such as ``Hint-Before-Solving'' and ``Progressive-Hint Prompting''.
    \item Activate context-aware hint production during task execution to dynamically guide the completion of generated plans. Use zero-shot CoT reasoning mantricas bolstered by training within diverse auxiliary datasets to continually support effective plan variations.
\end{itemize}

\textbf{3. Iterative Self-Training with Abstract Feedback Loop:}
\begin{itemize}[leftmargin=*, nosep]
    \item Establish a systematic process wherein models receive and evaluate abstract rationale-based feedback post-task execution to refine their initial planning attempts.
    \item Create mechanisms for trackable decision trees and failure resolutions allowing LLM calibrations through reinforcement with generated plan rationales, systematically measured within meta-learning validation processes.
\end{itemize}

\textbf{4. Adaptive Plan Specah Quality Assurance System:}
\begin{itemize}[leftmargin=*, nosep]
    \item Implement advanced sensing techniques to enable seamless decision-making regarding both the breadth (variety of request paths) and depth (granularity of step detail) of plans.
    \item Apply reinforcement learning frameworks such as PPO augmented clearly by detailed success outcomes linked closely with baseline predictions to elevate derived qualitative outputs.
\end{itemize}

\textbf{5. Integration of Direct Preference Optimization:}
\begin{itemize}[leftmargin=*, nosep]
    \item Innovatively employ Direct Preference Optimization (DPO) to create an orderly reward structure, which appropriately crafts LN complex outputs according to inherent internal merit transparency.
    \item Platform parallel functionality with measures from constructs like STaR experience loops to individually assay sourced plan evaluations validating emergence effects.
\end{itemize}
\end{baselinebox}
\caption{Generated \textbf{Method} by AI-Scientist-V2 for the same research problem.}
\label{fig:ai-scientist-v2-method}
\end{figure*}

\begin{figure*}[t]
\begin{baselinebox}{Method (Claude-3.5-Sonnet)}
\textbf{Hierarchical Planning with Adaptive Self-Refinement (HiPAR)}

We present Hierarchical Planning with Adaptive Self-Refinement (HiPAR), a framework that enables language models to develop increasingly sophisticated problem-solving capabilities through iterative plan generation and refinement. The key insight of our approach is to decompose complex reasoning tasks into a hierarchical planning phase followed by guided solution generation, while incorporating adaptive self-improvement mechanisms.

\textbf{Overview.} HiPAR consists of three main components: (1) A hierarchical planner that generates structured solution outlines at multiple levels of abstraction; (2) An execution module that converts plans into detailed solutions; (3) An adaptive refinement mechanism that improves both planning and execution through self-training.

\textbf{Hierarchical Planning.} Given an input problem $x$, the planner generates a hierarchical solution structure: $H(x) = \{P_1, P_2, ..., P_L\}$, where $P_i$ represents the plan at level $i$, with $P_1$ being the highest-level strategic outline and $P_L$ being the most detailed tactical plan. Each level $P_i$ consists of a sequence of plan steps: $P_i = \{p_{i,1}, p_{i,2}, ..., p_{i,n_i}\}$. The planning process proceeds top-down, with each level providing increasingly specific guidance for the levels below. We implement this using a novel recursive prompting strategy.

\textbf{Plan-Guided Solution Generation.} The execution module converts the hierarchical plan into a detailed solution by following a novel attention-guided generation process. Given the full plan hierarchy $H(x)$, we compute plan-aware attention weights:
$\alpha_{t,i,j} = \text{softmax}(f_\theta(h_t, p_{i,j}))$,
where $h_t$ is the hidden state at generation step $t$, and $f_\theta$ is a learnable scoring function. These weights determine how much each plan element influences the generation:
$c_t = \sum_{i=1}^L \sum_{j=1}^{n_i} \alpha_{t,i,j} g_\phi(p_{i,j})$.
The context vector $c_t$ is then used to guide the next token prediction:
$P(y_t|y_{<t}, x, H) = \text{softmax}(W[h_t; c_t])$.

\textbf{Adaptive Self-Refinement.} To enable continuous improvement, we introduce an adaptive self-refinement mechanism inspired by STaR but extended to handle hierarchical plans. For each training example $(x, y)$: (1) Generate initial hierarchical plan $H_0(x)$ and solution $\hat{y}_0$; (2) If $\hat{y}_0$ is incorrect, generate refined plan $H_1(x)$ conditioned on correct solution $y$; (3) Fine-tune both planner and execution module on successful $(H_1, y)$ pairs.

The refinement objective combines plan quality and solution accuracy:
$\mathcal{L} = \mathcal{L}_{\text{plan}}(H_1, H^*) + \lambda\mathcal{L}_{\text{exec}}(\hat{y}, y)$,
where $H^*$ represents expert-created plans (when available) and $\lambda$ balances the objectives.

\textbf{Training Algorithm.} The full training procedure alternates between: (1) Collecting experiences through hierarchical planning and execution; (2) Filtering successful planning traces; (3) Updating model parameters using filtered examples. We employ a novel curriculum that gradually increases plan complexity:
$L_t = \min(L_{\text{max}}, \lceil t/T \cdot L_{\text{max}} \rceil)$,
where $t$ is the current training step and $T$ is a schedule parameter.

This approach enables the model to progressively learn more sophisticated planning and problem-solving strategies while maintaining high solution quality through the adaptive refinement mechanism.
\end{baselinebox}
\caption{Generated \textbf{Method} by Claude-3.5-Sonnet for the same research problem.}
\label{fig:claude-sonnet-method}
\end{figure*}

\section{Analysis of Internalized Reasoning Behaviors}
\label{app:behavior_patterns}

Beyond the single end-to-end example in Appendix~\ref{app:qualitative_examples}, we further examine \textit{what kinds of reasoning behaviors} MoRI exhibits across diverse research contexts. By inspecting generated CoTs across the test set, we identify three recurring and qualitatively distinct behavior patterns: (A) \textbf{goal decomposition with constraint enumeration}, (B) \textbf{hypothesize--critique--revise loops}, and (C) \textbf{paradigm questioning via reverse framing}. Representative excerpts are shown in Figures~\ref{fig:behavior-a}--\ref{fig:behavior-c}.

These patterns collectively indicate that MoRI does not collapse to a single stylistic template. Instead, it adapts its reasoning strategy to the structure of the problem: structured engineering-oriented problems elicit decomposition, ill-specified problems elicit self-critique, and problems dominated by a methodological convention elicit paradigm questioning. This behavioral diversity provides additional qualitative evidence for \textbf{RQ3}, complementing the training-dynamics analysis in Section~\ref{subsec:ablation}.

\begin{figure*}[t]
\begin{reasoningbox}
\textbf{Behavior A --- Goal Decomposition with Constraint Enumeration.} The model first articulates the design target, enumerates the properties a valid solution must satisfy, and only then narrows toward a concrete mechanism.

\medskip

\insight{My goal is to create a method that not only averages model parameters but also incorporates historical model information} to stabilize the aggregation process. [\ldots]

\medskip

\insight{My solution should therefore involve:}
\begin{enumerate}[leftmargin=1.5em, noitemsep, topsep=2pt]
    \item A distance-based mechanism to assess how far a past model is from current knowledge.
    \item A weighted aggregation strategy where distant models are assigned smaller weights, favoring recent but informative snapshots.
\end{enumerate}

\medskip

\insight{The method must be lightweight --- no auxiliary tasks or new data sets --- merely using the existing local model parameters.} [\ldots]

\medskip

\textit{Thinking further}\ldots \insight{To avoid costly pairwise comparisons, the distance metric should be computed efficiently} in a way that scales with the number of participating models. [\ldots]
\end{reasoningbox}
\caption{\textbf{Behavior A: Goal Decomposition with Constraint Enumeration.} The model declares an objective, enumerates required properties, and narrows toward a concrete mechanism in a top-down manner.}
\label{fig:behavior-a}
\end{figure*}

\begin{figure*}[t]
\begin{reasoningbox}
\textbf{Behavior B --- Hypothesize--Critique--Revise Loop.} The model proposes candidate designs, identifies their failure modes on its own, and iteratively refines its hypothesis rather than committing to the first plausible answer.

\medskip

\insight{So here's an idea: use trees with a fixed branching factor and depth.} Each node's label is determined by the structure of its subtree. [\ldots]

\medskip

\insight{Ah! Here's the core insight: in a tree or a ring, the parity of the depth can be determined by just knowing} local neighborhood information. [\ldots]

\medskip

\insight{So that won't work.} We need something where each node's label depends on a signal that cannot be inferred from short-range structure alone. [\ldots]

\medskip

\insight{Wait a minute, maybe using a tree with a fixed root and levels} --- but to enforce that labels depend on some genuinely distant signal, we need an additional constraint. [\ldots]
\end{reasoningbox}
\caption{\textbf{Behavior B: Hypothesize--Critique--Revise Loop.} The model actively discovers shortcomings in its own proposals and iterates, demonstrating self-verification rather than single-shot generation.}
\label{fig:behavior-b}
\end{figure*}

\begin{figure*}[t]
\begin{reasoningbox}
\textbf{Behavior C --- Paradigm Questioning via Reverse Framing.} Rather than follow the dominant methodological convention for a given task, the model first interrogates \textit{why} that convention is used and identifies what the literature is collectively missing.

\medskip

\insight{My initial thought is: Why not flip the script?} Instead of building complex, fragile models that assume fixed dependency structures, we could start from a fundamentally different assumption. [\ldots]

\medskip

\insight{So, what's missing? The existing literature treats dependencies in a static, all-or-nothing way} --- either they are modeled monolithically or ignored entirely. Neither regime matches the underlying data-generating process. [\ldots]

\medskip

\insight{The key insight: dynamic structural selection should be local in both space and time.} A globally optimized dependency structure cannot capture phenomena that shift over time, while a fully localized view discards useful long-horizon signal. [\ldots]
\end{reasoningbox}
\caption{\textbf{Behavior C: Paradigm Questioning via Reverse Framing.} The model challenges the default methodological stance of the field before committing to a solution direction, enabling non-obvious research angles.}
\label{fig:behavior-c}
\end{figure*}

\end{document}